\documentclass[graybox]{svmult}    

\usepackage{mathptmx}       
\usepackage{helvet}         
\usepackage{courier}        
\usepackage{type1cm}        
\usepackage{makeidx}         
\usepackage{graphicx}        
\usepackage{multicol}        
\usepackage[bottom]{footmisc}
\usepackage{booktabs}

\makeindex             

\usepackage{geometry}
\usepackage[utf8]{inputenc} 
\usepackage[T1]{fontenc}    
\usepackage{hyperref}       
\usepackage{url}
\makeatletter
\g@addto@macro{\UrlBreaks}{\UrlOrds}
\makeatother

\usepackage{booktabs}       
\usepackage{amsfonts}       
\usepackage{nicefrac}       
\usepackage{microtype}      
\usepackage{tabularx}
\usepackage{caption}
\usepackage{mdwlist}
\usepackage{amssymb}
\usepackage{diagbox}
\usepackage{graphicx}
\usepackage{wrapfig}
\usepackage{listings}
\usepackage{color}
\usepackage{xcolor}
\usepackage{nicefrac}
\usepackage{subcaption}
\usepackage{amsmath}
\usepackage{multirow}

\usepackage{algorithm}
\usepackage{algorithmicx}
\usepackage{algpseudocode}
\usepackage{varwidth}

\usepackage{tikz}

\global\long\def\E{\mathbb{E}}

\usepackage{amsmath}
\usepackage{graphicx}

\usepackage{todonotes}
\usepackage{hyperref}
\usepackage{float}
\usepackage{amssymb}
\usepackage{amsmath}
\usepackage{comment}
\usepackage[normalem]{ulem}

\usepackage[T1]{fontenc}
\usepackage[utf8]{inputenc}
\usepackage[T1]{fontenc}
\usepackage[utf8]{inputenc}

\usepackage{caption}

\usepackage{titlesec} 

\usepackage{verbatim}
 \usetikzlibrary{decorations.pathreplacing}

\newcommand{\citep}[1]{\cite{#1}} 

\makeatletter
\newcommand{\sectionauthor}[1]{%
  {\parindent0pt\vspace*{-12pt}%
  \linespread{1.1}#1%
  \par\nobreak\vspace*{15pt}}
  \@afterheading%
}
\makeatother

\graphicspath{
{img/}
}
\begin{document}
\title*{Artificial Intelligence for Prosthetics --- challenge solutions}

\author{{\L}ukasz Kidzi\'nski, Carmichael Ong, Sharada Prasanna Mohanty, Jennifer Hicks, Sean F. Carroll,\
Bo Zhou, Hongsheng Zeng, Fan Wang, Rongzhong Lian, Hao Tian,\
Wojciech Jaśkowski, Garrett Andersen, Odd Rune Lykkebø, Nihat Engin Toklu, Pranav Shyam, Rupesh Kumar Srivastava,\
Sergey Kolesnikov, Oleksii Hrinchuk, Anton Pechenko,\
Mattias Ljungström,\
Zhen Wang, Xu Hu, Zehong Hu, Minghui Qiu, Jun Huang,\
Aleksei Shpilman, Ivan Sosin, Oleg Svidchenko, Aleksandra Malysheva, Daniel Kudenko,\
Lance Rane,\
Aditya Bhatt,\
Zhengfei Wang, Penghui Qi, Zeyang Yu, Peng Peng, Quan Yuan, Wenxin Li,\
Yunsheng Tian, Ruihan Yang, Pingchuan Ma,\
Shauharda Khadka, Somdeb Majumdar, Zach Dwiel, Yinyin Liu, Evren Tumer,\
Jeremy Watson,\
Marcel Salathé, Sergey Levine, Scott Delp}


\maketitle
\abstract{In the NeurIPS 2018 \textit{Artificial Intelligence for Prosthetics} challenge, participants were tasked with building a controller for a musculoskeletal model with a goal of matching a given time-varying velocity vector. Top participants were invited to describe their algorithms. In this work, we describe the challenge and present thirteen solutions that used deep reinforcement learning approaches. Many solutions use similar relaxations and heuristics, such as reward shaping, frame skipping, discretization of the action space, symmetry, and policy blending. However, each team implemented different modifications of the known algorithms by, for example, dividing the task into subtasks, learning low-level control, or by incorporating expert knowledge and using imitation learning.}


\section{Introduction}

Recent advancements in material science and device technology have increased interest in creating prosthetics for improving human movement. Designing these devices, however, is difficult as it is costly and time-consuming to iterate through many designs. This is further complicated by the large variability in response among many individuals. One key reason for this is that our understanding of the interactions between humans and prostheses is not well-understood, which limits our ability to predict how a human will adapt his or her movement. Physics-based, biomechanical simulations are well-positioned to advance this field as it allows for many experiments to be run at low cost. Recent developments in using reinforcement learning techniques to train realistic, biomechanical models will be key in increasing our understanding of the human-prosthesis interaction, which will help to accelerate development of this field.

\begin{figure}[ht!]
\centering
\includegraphics[width=0.72	\linewidth]{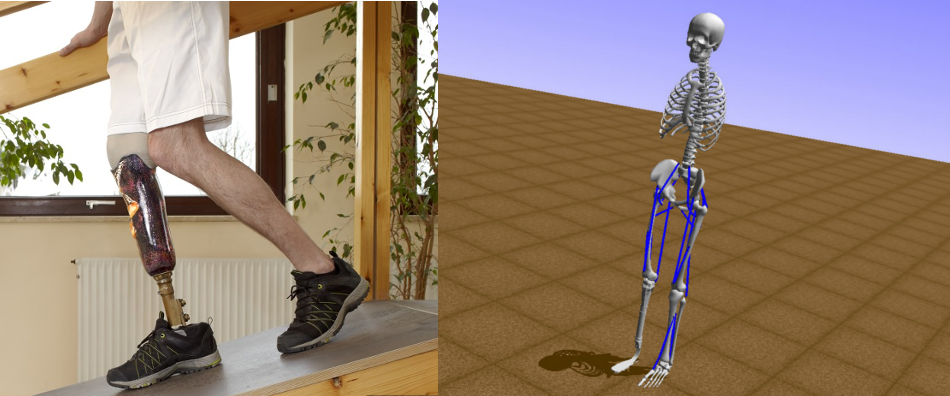}
\caption{A patient with a prosthetic leg (left); musculoskeletal simulation of a patient with a prosthetic leg as modeled in Stanford's OpenSim software (right).}
\label{fig:running}
\end{figure}

In this competition, participants were tasked with developing a controller to enable a physiologically-based human model with a prosthetic leg to move at a specified direction and speed. Participants were provided with a human musculoskeletal model and a physics-based simulation environment (OpenSim \cite{delp2007opensim,seth2018opensim}) in which they synthesized physically and physiologically accurate motion (Figure \ref{fig:running}). Entrants were scored based on how well the model moved according to the requested speed and direction of walking. We provided competitors with a parameterized training environment to help build the controllers, and competitors' scores were based on a final environment with unknown parameters.


This competition advanced and popularized an important class of reinforcement learning problems, characterized by a large set of output parameters (human muscle controls) and a comparatively small dimensionality of the input (state of a dynamic system). Our challenge attracted over 425 teams from the computer science, biomechanics, and neuroscience communities, submitting 4,575 solutions. Algorithms developed in this complex biomechanical environment generalize to other reinforcement learning settings with highly-dimensional decisions, such as robotics, multivariate decision making (corporate decisions, drug quantities), and the stock exchange.

In this introduction, we first discuss state-of-the-art research in motor control modeling and simulations as a tool for solving problems in biomechanics (Section \ref{ss:scope}). Next, we specify the details of the task and performance metrics used in the challenge (Section \ref{ss:task}). Finally, we discuss results of the challenge and provide a summary of the common strategies that teams used to be successful in the challenge (Section \ref{ss:solutions}). In the following sections, top teams describe their approaches in more detail.

\subsection{Background and scope}\label{ss:scope}


Using biomechanical simulations to analyze experimental data has led to novel insights about human-device interaction. For example, one group used simulations to study a device that decreased the force that the ankle needed to produce during hopping but, paradoxically, did not reduce energy expenditure. Simulations that included a sophisticated model of muscle-tendon dynamics revealed that this paradox occurred because the muscles were at a length that was less efficient for force production \cite{farris2014exo}. Another study used simulations of running to calculate ideal device torques needed to reduce energy expenditure during running\cite{uchida2016device}, and insights gained from that study were used to decrease the metabolic cost in experimental studies \cite{lee2017exosuit}

A limitation of the previous studies, however, is that they used the kinematics from the experiments as a constraint. Since assistive devices may greatly change one's kinematics, these analyses cannot predict the neural and kinematic adaptations induced by these devices on the human. Instead, we require a framework that can generate simulations of gait that can adapt realistically to various perturbations. This framework, for instance, would help us understand the complex motor control and adaptations necessary to control both a healthy human and one with a prosthetic.


Recent advances in reinforcement learning, biomechanics, and neuroscience enable us to build a framework that tackles these limitations. The biomechanics community has developed models and controllers for different movement tasks, two of which are particularly relevant for this challenge. The first study developed a model and controller that could walk, turn, and avoid obstacles \cite{song2015neural}. The second study generated simulated walking patterns at a wide range of target speeds that reproduced many experimental measures, including energy expenditure \cite{ong2017walking}. These controllers, however, are limited to generating walking and running and need domain expertise to design.

Modern reinforcement learning techniques have been used recently to train more general controllers in locomotion. Controllers generated using these techniques have the advantage that, compared to the gait controllers previously described, less user input is needed to hand-tune the controllers, and they are more flexible in their ability to learn additional, novel tasks. For example, reinforcement learning has been used to train controllers for locomotion of complicated humanoid models \cite{lillicrap2015continuous,schulman2015trust}. Although these methods find solutions without domain specific knowledge, the resulting motions are not realistic, possibly because these models do not use biologically accurate actuators.

Through the challenge, we aimed to investigate if deep reinforcement learning methods can yield more realistic results with biologically accurate models of the human musculoskeletal system. We designed the challenge so that it stimulates research at the intersection of reinforcement learning, biomechanics, and neuroscience, encouraging development of methods appropriate for environments characterized by the following: 1) a high-dimensional action space, 2) a complex biological system, including delayed actuation and complex muscle-tendon interactions, 3) a need for a flexible controller for an unseen environment, and 4) an abundance of experimental data that can be leveraged to speed up the training process.

\subsection{Task}\label{ss:task}

Competitors were tasked with building a real-time controller for a simulated agent to walk or run at a requested speed and direction. The task was designed in a typical reinforcement learning setting \cite{Sutton1999} in which an agent (musculoskeletal model) interacts with an environment (physics simulator) by taking actions (muscle excitations) based on observations (a function of the internal state of the model) in order to maximize the reward.

\textbf{Agent and observations.} The simulated agent was a musculoskeletal model of a human with a prosthetic leg. The model of the agent included $19$ muscles to control $14$ degrees-of-freedom (DOF). At every iteration the agent received the current observed state, a vector of consisting of $406$ values: ground reaction forces, muscle activities, muscle fiber lengths, muscle velocities, tendon forces, and positions, velocities, and accelerations of joint angles and body segments.

Compared to the original model from \cite{ong2017walking} we allowed the hip to abduct and adduct, and the pelvis to move freely in space. To control the extra hip degrees of freedom, we added a hip abductor and adductor to each leg, which added 4 muscles total. The prosthesis replaced the right tibia and foot, and we removed the 3 muscles that cross the ankle for that leg. Table \ref{tbl:action} lists all muscles in the model. 

\begin{table}[h!]
\setlength\tabcolsep{0.25cm}
\setlength{\extrarowheight}{.2em}
\centering
\begin{tabular}{r l l l } 
Name & Side	& Description &	Primary function(s)\\
 \hline
abd & both & Hip abductors & Hip abduction (away from body’s vertical midline)\\
add & both & Hip adductors & Hip adduction (toward body’s vertical midline)\\
bifemsh & both & Short head of the biceps femoris &	Knee flexion\\
gastroc & left & Gastrocnemius & Knee flexion and ankle extension (plantarflexion)\\
glut\_max & both &	Gluteus maximus & Hip extension\\
hamstrings & both &	Biarticular hamstrings & Hip extension and knee flexion\\
iliopsoas & both &	Iliopsoas &	Hip flexion\\
rect\_fem &	both & Rectus femoris &	Hip flexion and knee extension\\
soleus & left &	Soleus & Ankle extension (plantarflexion)\\
tib\_ant & left & Tibialis anterior & Ankle flexion (dorsiflexion)\\
vasti & both & Vasti & Knee extension\\
\end{tabular}
\caption{A list of muscles that describe the action space in our physics environment. Note that some muscles are missing in the amputated leg (right).}
\label{tbl:action}
\end{table}

\textbf{Actions and environment dynamics.} Based on the observation vector of internal states, each participant's controller would output a vector of muscle excitations (see Table \ref{tbl:action} for the list of all muscles). The physics simulator, OpenSim, calculated muscle activations from excitations using first-order dynamics equations. Muscle activations generate movement as a function of muscle properties such as strength, and muscle states such as current length, velocity, and moment arm. An overall estimate of muscle effort was calculated using the sum of muscle activations squared, a commonly used metric in biomechanical studies \cite{crowninshield1981, thelen2003cmc}. Participants were evaluated by overall muscle effort and the distance between the requested and observed velocities.


\textbf{Reward function and evaluation.} Submissions were evaluated automatically. In Round 1, participants interacted directly with a remote environment. The overall goal of this round was to generate controls such that the model would move forward at 3 m/s. The total reward was calculated as

$$ \sum_{t=1}^{T} 9 - |v_x(s_t) - 3|^2, $$

where \(s_t\) is the state of the model at time \(t\), \(v_x(s)\) is the horizontal velocity vector of the pelvis in the state \(s\), and \(s_t = M(s_{t-1}, a(s_{t_1}))\), i.e. states follow the simulation given by model \(M\). $T$ is the episode termination time step, which is equal to $1000$ if the model did not fall for the full 10 second duration, or is equal to the first time point when that the pelvis of the model falls below $0.6$ m to penalize falling.

In Round 2, in order to mitigate the risk of overfitting, participants submitted Docker containers so that they could not infer the specifics of the environment by interacting with it. The objective was to move the model according to requested speeds and directions. This objective was measured as a distance between the requested and observed velocity vector.

The requested velocity vector varied during the simulation. We commanded approximately $3$ changes in direction and speed during the simulation. More precisely, let $q_0 = 1.25,r_0 = 1$ and let $N_t$ be a Poisson process with $\lambda = 200$. We define
\[
    q_t = q_{t-1} + \mathbf{1}_{(N_t \neq N_{t-1})} u_{1,t} \text{ and }
    r_t = r_{t-1} + \mathbf{1}_{(N_t \neq N_{t-1})} u_{2,t},
\]
where $\mathbf{1}_{(A)}$ is the indicator function of the event $A$ (here, a jump in the Poisson process). We define $u_{1,t} \sim \mathcal{U}([-0.5,0.5])$ and $u_{2,t} \sim \mathcal{U}([-\pi/8,\pi/8])$, where $\mathcal{U}([a,b])$ denotes a uniform distribution on $[a,b]$ interval.

In order to keep the locomotion as natural as possible, we also added a penalties for overall muscle effort as validated in previous work \cite{crowninshield1981, thelen2003cmc}. The final reward function took form

\[
\sum_{t=1}^{T} \left( 10 - |v_x(s_t) - w_{t,x} |^2 - |v_z(s_t) - w_{t,z} |^2 - 0.001\sum_{i=1}^{d}a_{t,i}^2 \right),
\]
where $w_{t,x}$ and $w_{t,z}$ correspond to $q_t$ and $r_t$ expressed in Cartesian coordinates,  $T$ is termination step of the episode as described previously, $a_{t,i}$ is the activation of muscle $i$ at time $t$, and $d$ is the number of muscles.


Since the environment is subject to random noise, submissions were tested over ten trials and the final score was the average from these trials.

\subsection{Solutions}\label{ss:solutions}

Our challenge attracted 425 teams who submitted 4,575 solutions. The top 50 teams from Round~1 qualified for Round~2. In Table \ref{tbl:leaderboard} we list the top teams from Round~2. Detailed descriptions from each team are given in the subsequent sections of this article. Teams that achieved 1st through 10th place described their solutions in sections 2 through 11. Two other teams submitted their solutions as well (Sections 12 and 13).

\begin{table}[h!]
\setlength\tabcolsep{0.25cm}
\setlength{\extrarowheight}{.2em}
\centering
\begin{tabular}{r | l l l l l } 
  & Team & Score & \# entries & Base algorithm\\
 \hline
 1 & Firework       & 9981 & 10 & DDPG \\ 
 2 & NNAISENSE      & 9950 & 10 & PPO  \\
 3 & Jolly Roger    & 9947 & 10 & DDPG  \\
 4 & Mattias       & 9939 & 10 & DDPG  \\
 5 & ItsHighNoonBangBangBang         & 9901 & 3  & DDPG  \\
 6 & jbr            & 9865 & 9  & DDPG  \\ 
 7 & Lance          & 9853 & 4  & PPO \\
 8 & AdityaBee      & 9852 & 10 & DDPG  \\
 9 & wangzhengfei   & 9813 & 10 & PPO  \\
 10 & Rukia         & 9809 & 10 & PPO  \\
\end{tabular}
\caption{Final leaderboard (Round 2).}
\label{tbl:leaderboard}
\end{table}

In this section we highlight similarities and differences in the approaches taken by the teams. Of particular note was the amount of compute resources used by the top participants. Among the top ten submissions, the highest amount of resources reported for training the top model was 130,000 CPU-hours, while the most compute-efficient solution leveraged experimental data and imitation learning (Section~\ref{s:lance}) and took only 100 CPU-hours to achieve 7th place in the challenge (CPU-hours were self-reported). Even though usage of experimental data was allowed in this challenge, most participants did not use it and focused only on reinforcement learning with a random starting point. While such methods are robust, they require very large compute resources. 

While most of the teams used variations of well-established algorithms such as DDPG and PPO, each team used a combination of other strategies to improve performance. We identify some of the key strategies used by teams below.\\
\ \newline
\noindent\textbf{Leveraging the model.} Participants used various methods to encourage the model to move in a realistic way based on observing how humans walk. This yielded good results, likely due to the realistic underlying physics and biomechanical models. Specific strategies to leverage the model include the following:
\begin{itemize}
    \item Reward shaping: Participants modified the reward for training in such a way that it still makes the model train faster for the actual initial reward. (see, for example, Sections \ref{sss:reward-jolly}, \ref{sss:reward-mattias}, or \ref{sss:reward-itshigh}).
    \item Feature engineering: Some of the information in the state vector might add little value to the controller, while other information can give a stronger signal if a non-linear mapping based on expert knowledge is applied first (see, for example, Sections \ref{sss:reward-jolly}, \ref{sss:observation-mattias}, or \ref{sss:observation-jbr}). Interestingly, one team achieved a high score without feature engineering (Section \ref{sss:large-space}).
    \item Human-in-the-loop optimization: Some teams first trained a batch of agents, then hand picked a few agents that performed well for further training (Section \ref{sss:human-in-the-loop}).
    \item Imitation learning: One solution used experimental data to quickly find an initial solution and to guide the controller towards typical human movement patterns. This resulted in training that was quicker by a few orders of magnitude (Section \ref{s:lance}).
\end{itemize}
\ \newline
\noindent\textbf{Speeding up exploration.} In the early phase of training, participants reduced the search space or modified the environment to speed up exploration using the following techniques:
\begin{itemize}
    \item Frameskip: Instead of sending signals every $1/100$ of a second (i.e., each frame), participants sent the same control for, for example, 5 frames. Most of the teams used some variation of this technique (see, for example, Section \ref{sss:observation-mattias})
    \item Sample efficient algorithms: all of the top teams used algorithms that are known to be sample-efficient, such as PPO and DDPG. 
    \item Exploration noise: Two main exploration strategies involved adding Gaussian or Ornstein--Uhlenbeck noise to actions (see Section \ref{sss:parameter-jolly}) or parameter noise in the policy (see Section \ref{s:nnaisense} or \ref{sss:parameter-itshigh}).
    \item Binary actions: Some participants only used muscle excitations of exactly 0 or 1 instead of values in the interval $[0,1]$ (``bang-bang'' control) to reduce the search space (Section \ref{sss:large-space}).
    \item Time horizon correction: An abrupt end of the episode due to a time limit can potentially mislead the agent. To correct for this effect, some teams used an estimate of the value behind the horizon from the value function (see Section \ref{sss:time-horizon}).
    \item Concatenating actions: In order to embed history in the observation, some teams concatenated observations before feeding them to the policy (Section \ref{sss:concatenation}).
    \item Curriculum learning: Since learning the entire task from scratch is difficult, it might be advantageous to learn low-level tasks firsts (e.g., bending the knee) and then learn high-level tasks (e.g., coordinating muscles to swing a leg) (Section \ref{ss:methodology})
    \item Transfer learning: One can consider walking at different speeds as different subtasks of the challenge. These subtasks may share control structure, so the model trained for walking at 1.25 m/s may be retrained for walking at 1.5 m/s etc. (Section \ref{ss:methodology})
\end{itemize}
\ \newline
\noindent\textbf{Speeding up simulations.} Physics simulations of muscles and ground reaction forces are computationally intensive. Participants used the following techniques to mitigate this issue: 
\begin{itemize}
    \item Parallelization: Participants ran agents on multiple CPUs. A master node, typically with a GPU, was collecting these experiences and updating weights of policies. This strategy was indispensable for success and it was used by all teams (see, for example, Sections \ref{sss:distributed-jolly}, \ref{sss:distributed-apex}, \ref{sss:distributed-jbr} or  \ref{sss:distributed-shawk})
    \item Reduced accuracy: In OpenSim, the accuracy of the integrator is parameterized and can be manually set before the simulation. In the early stage of training, participants reduced accuracy to speed up simulations to train their models more quickly. Later, they fine-tuned the model by switching the accuracy to the same one used for the competition \cite{kidzinski2018learning}.
\end{itemize}
\ \newline
\noindent\textbf{Fine-tuning.} A common statistical technique for increasing the accuracy of models is to output a weighted sum of multiple predictions. This technique also applies to policies in reinforcement learning, and many teams used some variation of this approach: ensemble of different checkpoints of models (Section \ref{sss:ensemble}), training multiple agents simultaneously (Section \ref{s:mattias}), or training agents with different seeds (Section \ref{s:jbr}).\\
\ \newline
While this list covers many of the commonly used strategies, a more detailed discussion of each team's approach is detailed in the following sections.



\section{Efficient and Robust Learning on Elaborated Gaits with Curriculum Learning}\label{s:BaiDu_NLP}
\sectionauthor{Bo Zhou, Hongsheng Zeng, Fan Wang, Rongzhong Lian, Hao Tian}

We introduce a new framework for learning complex gaits with musculoskeletal models. Our framework combines Reinforcement Learning with Curriculum Learning \cite{Yoshua2009Curriculum}. We used Deep Determinsitc Policy Gradient (DDPG) \cite{lillicrap2015continuous} which is driven by the external control command. We accelerated the learning process with large-scale distributional training and bootstrapped deep exploration \cite{Ian2016Deep} paradigm. As a result, our approach\footnote{Find open-source code at: https://github.com/PaddlePaddle/PARL} won the NeurIPS 2018: AI for Prosthetics competition, scoring more than 30 points than the second placed solution.

\subsection{Challenges}
Compared with the 2017 Learning To Run competition, there are several changes in 2018 AI for Prosthetics competition. Firstly, the model is no longer restricted to 2D movement, but rather the model has 3D movement, including the lateral direction. Secondly, a prosthetic leg without muscles replaces the intact right leg. Thirdly, external random velocity command is provided, requiring the model to run at the specified direction and speed, instead of running as fast as possible. These changes raise a more functional challenge on human rehabilitation. We believe that there are several challenges in this problem. 

\begin{figure*}[ht!]%
       \centering
       \includegraphics[width=1\textwidth]{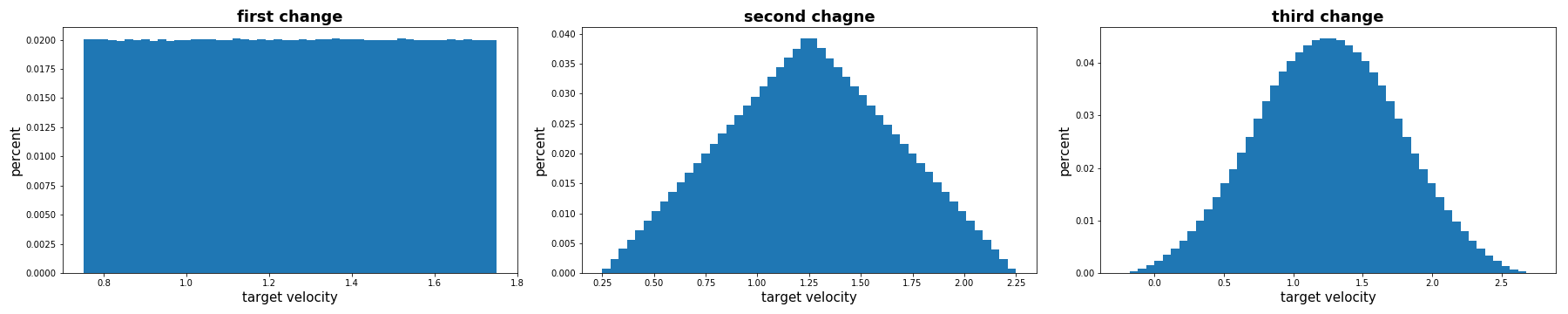}
       \caption{\textbf{Target velocity distribution.} Each distribution is based on $10^7$ samplings of the target velocity after each change.}
       \label{fig:velocity}
\end{figure*}

\textbf{High-Dimensional Non-Linear System.}
There are 185 dimension of observations, with 7 joints and 11 body parts in the whole system. The action space includes 19 continuous control signals for the muscles. Though the number of observation dimensions is not extremely large, the system is highly non-linear, and, furthermore, the action space is relatively large compared with many other problems. Moreover, as shown in Fig.~\ref{fig:velocity}, the agent is required to walk at different speeds and directions, which further expands the observation space and transition space. The curse of dimensionality \cite{Richard1961Adaptive} raises core issues of slow convergence, local optimum, and instability.

\textbf{Local optimum \& Stability.}
Though the local optimum problem is commonly faced in most dynamic systems, low-speed walking for the current model is especially problematic. According to Fig.~\ref{fig:velocity}, the required speed range is relatively low around 1.25m/s. If we try learning from scratch to achieve some specific speed, our early investigation revealed that the skeleton walks with a variety of gestures that result in pretty much the same performance in rewards. The agent either walks in a lateral direction (crab-like walking), bumping, or dragging one of its leg. While none of those walking gaits is natural, they are nearly indistinguishable in the rewards. However, although we found that those unrealistic gaits can reasonable generate static velocity walking, they perform very poorly with respect to stability. Transferring the model to other specified velocities becomes a problem, and the system is prone to fall, especially in the moment of switching velocities.

\subsection{Methodology}\label{ss:methodology}

To deal with the challenges mentioned above, we tried several main ideas. As there are a variety of basic RL algorithms to choose from, we chose DDPG \cite{lillicrap2015continuous} as DDPG is an efficient off-policy solver for continuous control. PPO, as an on-policy solver, often suffers from the problem of larger variance and lower efficiency. To further increase our efficiency, we applied Deep Exploration with multi-head bootstrapping \cite{Ian2016Deep}, which has been proven to converge much faster compared with $\epsilon$-greedy. In order to allow our policy to closely follow the velocity target, we inject the velocity as a feature to the policy and value network. At last, to address the core issue of local optimum, we applied curriculum learning to transfer efficient and stable gaits to various velocity range. 

\begin{figure}
\centering
\includegraphics[width=0.25\linewidth]{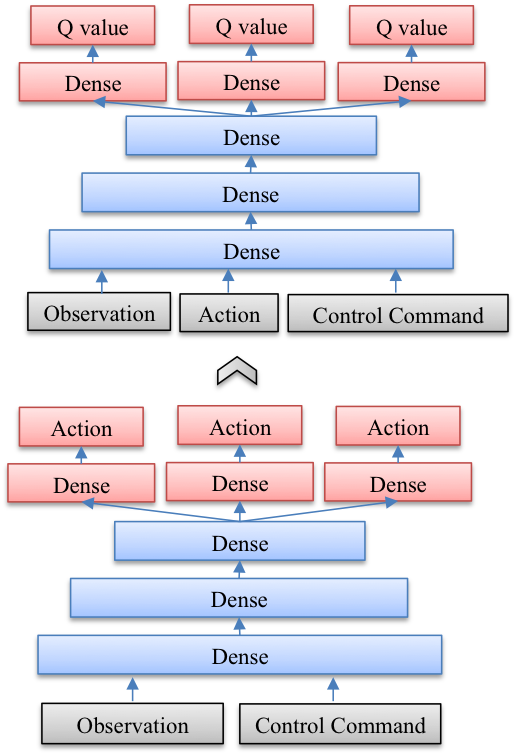}
\caption{Architecture of target-driven DDPG.}
\label{fig:target-driven}
\vspace*{-.2in}
\end{figure}

\textbf{Model architecture.} As shown in Fig.~\ref{fig:target-driven}, we used 4 fully connected layers for both actor network and critic network in the DDPG algorithm. Compared to general DDPG network architectures, it has two distinct features. We inject the target velocity from the bottom of both networks, as the value function needs to evaluate the current state based on target velocity, and the policy needs to take the corresponding action to reach the target velocity. This is similar to adding the target velocity as part of the observation. Though it introduces some noise when the velocity is switching, it benefits more by automatically sharing the knowledge of different velocities. 

We also use multiple heads for the value and policy network in our model. It is a similar architecture as deep exploration \cite{Ian2016Deep}, which simulates the ensemble of neural networks with lower cost by sharing the bottom layers. 

\textbf{Transfer Learning.} We propose that by sharing knowledge of walking or running at different speeds, the agent can learn more robust and efficient patterns of walking. We found in our experiment that the unstable gaits learned from scratch for low speed walking do not work well for high-speed running. We investigated on running as fast as possible instead of at a specified speed. We obtained an agent that can run very fast with reasonable and natural gaits just as humans. Starting with the trained policy for fast running, we switched the target to lower speed walking. This process resembles transfer learning, where we want the "knowledge" of gait to be kept but with a slower speed. Our fastest running has velocities over 4.0 m/s. We transferred the policy to 1.25 m/s, but it results in gestures that are still not natural enough and were prone to falling. Still, we made progress by transferring from a higher speed as the fall rate drops substantially.

\textbf{Curriculum Learning.} Curriculum learning \cite{Yoshua2009Curriculum} learns a difficult task progressively by artificially constructing a series of tasks which increase the difficulty level gradually. Recently it has been used to solve complex video game challenges\cite{Yuxin2017Training}. As the direct transfer of a higher speed running policy to lower speed did not work well, we devised 5 tasks to decrease the velocity linearly, with each task starting with the trained policy of the former one. At last, we have a policy running at target = 1.25m/s, with natural gaits that resembles a human being and low falling rate as well.

\textbf{Fine-tuning.} 
Based on the pretrained walking model targeted at 1.25 m/s, we fine-tune the model on the random velocity environment. Firstly, we try to force the policy to walk at 1.25 m/s, given any target velocity between -0.5 m/s and 3.0 m/s. This is to make a good start for other target velocities besides 1.2 m/s. We collect walking trajectories at 1.25 m/s, but change the features of target velocity and direction to a random value. We use the collected trajectories to re-train the policy with supervised learning. Secondly, we use the re-trained model as the start point, and fine-tune it in the randomized target velocity environments using target-driven DDPG, which gives our final policy.


\begin{figure}
\centering
\includegraphics[width=0.8\linewidth]{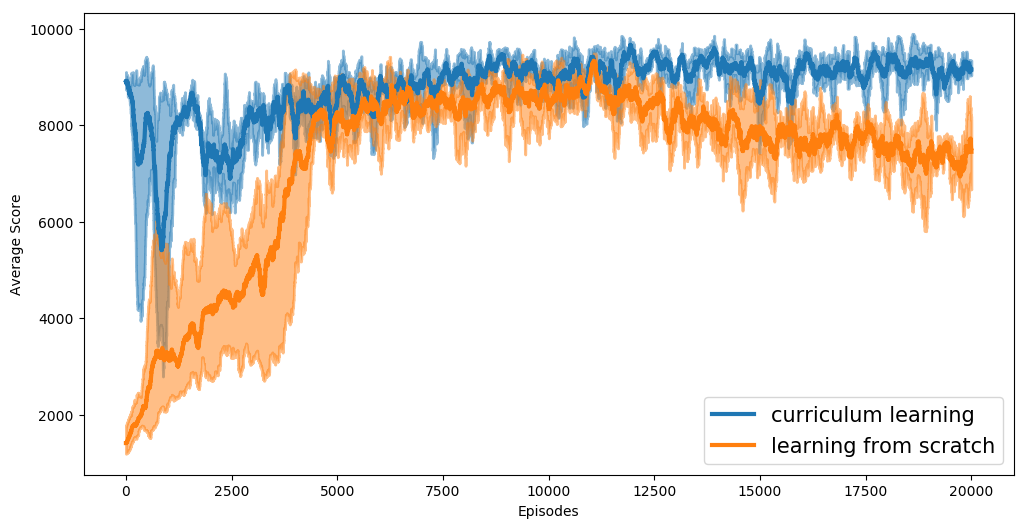}
\caption{Learning curves comparing curriculum-learning to learning from scratch. Average scores are computed from 50 episodes.}
\label{fig:curriculum-learning}
\end{figure}
 
\subsection{Experiments}

Our experiments compared curriculum learning with learning from scratch in the fine-tuning stage. We use the same model architecture for both experiments. For the actor model, we use tanh as activation function for each layer. For the critic model, selu \cite{klambauer2017self} is used as activation functions in each layer except the last layer. The discount factor for cumulative reward computation is 0.96. We also use the frame-skip trick, as each step of the agent corresponds to 4 simulation step in the environment with the same action. Twelve heads are used for bootstrapped deep exploration. This is decided by considering the trade-off between the diverse generalization of each head and computation cost in practice.

\begin{figure}
\centering
\includegraphics[width=0.8\linewidth]{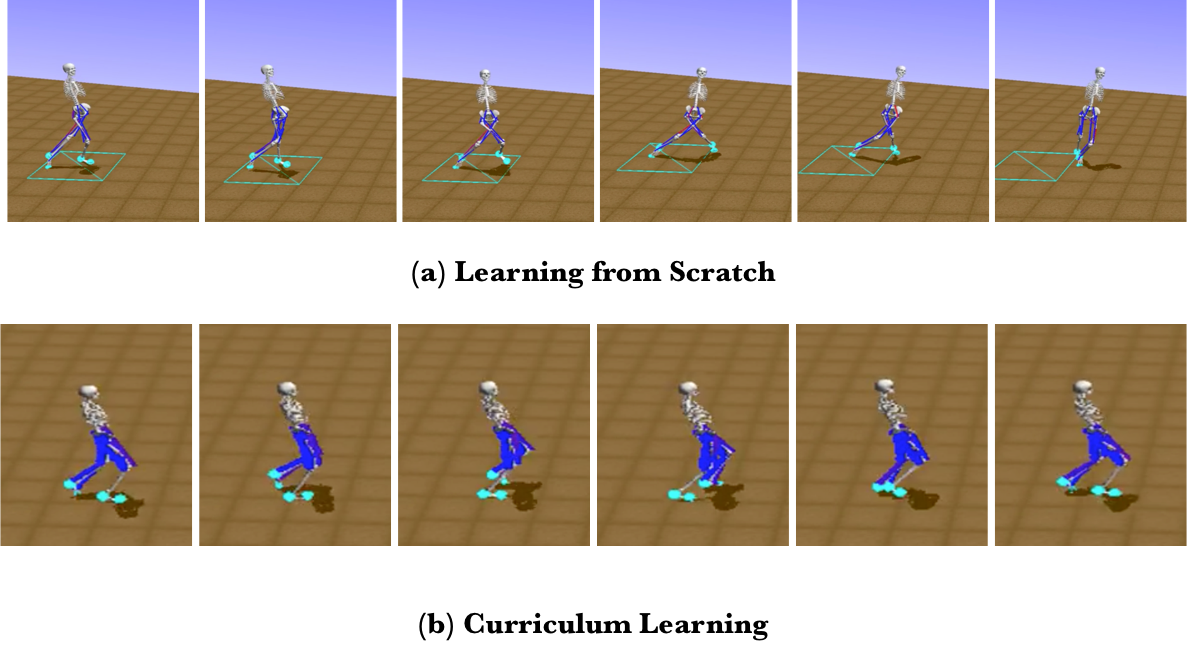}
\caption{Learned gaits. (a) The agent walks forward while heading at strange directions. (b) The skeleton walks naturally with small steps.}
\label{fig:learned-gaits}
\end{figure}

Figure~\ref{fig:curriculum-learning} shows the comparison of learning from scratch and starting from a policy learned with curriculum learning. Each curve is averaged on 3 independent experiments. Significant improvements on both performance and stability for the curriculum learning can be observed. Further investigating the walking gaits shows that curriculum learning has a more natural walking gesture, as shown in Figure~\ref{fig:learned-gaits}, while learning from scratch group results in crab-like walking.

\subsection{Discussion}
We have shown that curriculum learning is able to acquire sensible and stable gait. Firstly, we train the policy model to run as fast as possible to learn human-like running gestures.  We then diminish the target velocity gradually in later training. Finally, we fine-tune the policy in randomized velocity environment. Despite the state-of-the-art performance, there are still questions to be answered. Applying reinforcement learning in large and non-linear state and action spaces remains challenging. In this problem, we show that a sophisticated start policy is very important. However, the reason that running as fast as possible learns better gait is not yet fully answered. Moreover, the curriculum courses are hand-designed. Devising universal learning metrics for such problems can be very difficult. We are looking forward to further progresses in this area.

\section{Bang-Bang Control with Interactive Exploration to Walk with a Prosthetic}\label{s:nnaisense}
\sectionauthor{Wojciech Jaśkowski, Garrett Andersen, Odd Rune Lykkebø, Nihat Engin Toklu, Pranav Shyam, Rupesh Kumar Srivastava}

Following the success \cite{jaskowski2018rltorunfast} in the NeurIPS 17 ``Learning to Run'' challenge, the NNAISENSE Intelligent Automation team used a similar approach for the new NeurIPS 18 ``AI for Prosthetics'' contest. The algorithm involved using the Proximal Policy Optimization for learning a ``bang-bang'' control policy and human-assisted behavioural exploration. Other important techniques were i) a large number of features, ii) time-horizon correction, and, iii) parameter noise. Although the approach required a huge number of samples for fine-tuning the policy, the learning process was robust, which led NNAISENSE to win round 1 and to place second in round 2 of the competition.

\subsection{Methods}
\subsubsection{Policy Representation}\label{sss:large-space}
A stochastic policy was used. Both the policy function $\pi_\theta$, and the value function,
$V_\phi$, were implemented using feed-forward neural networks with two
hidden layers, with $256$ $\tanh$ units each.

The network input consisted of all $406$ features provided by the environment. The joint positions ($x$, and $z$'s) were made relative to the position of the pelvis. In addition, the coordinate system was rotated around the vertical axis to zero out the $z$ component of the target velocity vector. All the features were standardized with a running mean and variance.

Instead of the typical Gaussian policy, which gives samples in $[0,1]^d$, our network outputs a Bernoulli policy, which gives samples from $\{0,1\}^d$. Previously \cite{jaskowski2018rltorunfast}, it was found out that restricting control in this way leads to better results, presumably, due to reducing the policy space, more efficient exploration, and biasing the policy toward action sequences that are more likely to activate the muscle enough to actually generate movement.

To further improve exploration and reduce the search space, in the first part of  training, each action was executed for $4$ consecutive environment steps.

Our policy network utilized parameter noise \cite{fortunato2017noisy}, where the network weights are perturbed by Gaussian noise. Parameter noise was implemented slightly differently for on-policy and off-policy methods, interestingly, we found that our on-policy method benefited most from using the off-policy version of parameter noise.

\subsubsection{Policy Training}\label{sss:time-horizon}
The policy parameters $\theta$, $\phi$, and $\psi$ were learned with Proximal Policy Gradient (PPO, \cite{ppo}) with the Generalized Advantage Estimator (GAE, \cite{schulman2015high}) as the target for the advantage function.

A target advantage correction was applied in order to deal with the non-stationarity of the environment caused by the timestep limit. The correction, described in detail in \cite{jaskowski2018rltorunfast}, hides from the agent the termination of the episode that is caused by the timestep limit by making use of the value estimate. In result, it improves the precision of the value function, thus reducing the variance of the gradient estimator.

\subsubsection{Training Regime}\label{sss:human-in-the-loop}
The methodology applied to the NeurIPS competition task consisted of three stages: i) global initialization, ii) policy refinement, and iii) policy fine-tuning.

PPO is susceptible to local minima. Being an on-policy algorithm, every iteration improves the policy by a little bit. In result, it is unlikely to make large behavioural changes of the agent. Once the agent starts to exhibit a certain gait, PPO is unable to switch to a completely different way of walking later. To alleviate this problem, during the global initialization stage, $50$ runs were executed in parallel. After around $1000$ iterations, two gaits were selected based on their performance and behavioural dissimilarity to be improved in the subsequent stages.

The second stage, policy refinement, involved a larger number of samples per run and lasted until a convergence was observed. Afterwards, the steps per action parameter was reduced to the default $1$.

In the final stage, policy fine-tuning, all the exploration incentives were eventually turned off and the policy was specialized into two sub-policies, one for each task modes: i) \textit{ready-set-go}, used for the first $100$ timesteps; and ii) \textit{normal operation} for the rest of the episode. For the last day of the training, $576$ parallel workers were used (see Table~\ref{tab:nnhyperparams} for details).

\begin{table}[ht]
\centering
\caption{The hyper-parameters used and statistics used for subsequent learning stages.}
\label{tab:nnhyperparams}
\setlength\tabcolsep{0.25cm}
\setlength{\extrarowheight}{.1em}
\begin{tabular}{l c cc cc c}
                        \toprule
                         & \multicolumn{5}{c}{\textbf{Training stage}}
                         & \textbf{} \\
\cmidrule(l){2-6}
                         
                         & \textbf{Global initialization} &
                         \multicolumn{2}{c}{\textbf{Refinement}} & 
                         \multicolumn{2}{c}{\textbf{Specialization}} & 
                         
                         \textbf{Total} \\
                         \cmidrule(l){2-6}
                         
\textbf{}  &  \textbf{} & \textbf{I}    & \textbf{II}   & \textbf{I}      & \textbf{II}     &  \\
\cmidrule(l){1-7}
Parallel runs            & 50                             & 2                  & 2                  & 2                    & 2                    &                \\
Iterations $[\times1000]$              & 1.2                           & 1.3               & 1.7               & 1.6                 & 0.3                  & 6.2           \\
Episodes $[\times10^{5}]$                & 0.5                       & 7.8           & 2.6           & 2.8             & 3.6             & 17.3       \\
Steps $[\times10^{8}]$                   & 0.4                       & 7.8           & 2.6           & 2.8             & 3.6             & 17.1       \\
Training time $[\text{dyas}]$           & 3.8                          & 3.9              & 2.7              & 3.5                & 0.9                & 14.5         \\
Resources used $[\text{CPU hours}\times1000]$                & 36                       & 27           & 16           & 24             & 26             & 130      \\
\cmidrule(l){1-7}
Workers                  & 8                              & 144                & 144                & 144                  & 576                  &                \\
Steps per worker         & 1,024                          & 1,024              & 1,024              & 1,024                & 2,048                &                \\
Steps per action     & 4                                  & 4                  & 1                  & 1                    & 1                    &                \\
Entropy Coeff        & 0.01                 & 0.01        & 0.01               & 0.01                  &        0       & \\
Parameter noise               & yes                            & yes                & yes                & yes                  & no                   &                \\
Policy networks             & 1                             & 1                 & 1                 & 2                  & 2                  &                \\
\cmidrule(l){1-7}
PPO learning rate & \multicolumn{6}{c}{3$\times10^3$} \\
PPO clip parameter ($\epsilon$)    & \multicolumn{6}{c}{0.2}  \\
PPO batch size                  & \multicolumn{6}{c}{256}  \\
PPO optimizations per epoch        & \multicolumn{6}{c}{10}  \\
PPO input normalization clip   & \multicolumn{6}{c}{5 SD}  \\
PPO entropy coefficient        & \multicolumn{6}{c}{0}  \\
GAE $\lambda$ GAE               & \multicolumn{6}{c}{0.9}  \\
GAE $\gamma$ in GAE                & \multicolumn{6}{c}{0.99}  \\
\cmidrule(l){1-7}
\textbf{Final avg. score during training} & \textbf{9796}                  & \textbf{9922}      & \textbf{9941}      & \textbf{9943}            & \textbf{9952}        &                \\ 
\bottomrule
\end{tabular}
\end{table}

\subsubsection{Results}
The described training method resulted in two distinct gaits of similar average performance (see Table~\ref{tab:nnresults}). Both gaits have interesting characteristics. The slightly better policy (the ``Dancer''\footnote{\url{https://youtu.be/ckPSJYLAWy0}}), starts forward with his prosthetic leg, then turns around during his first few steps, and finally continues his walk backward from this time on. It seems the training found that walking backwards was a more efficient way to deal with the changes of the target velocity vector.

The other policy (the ``Jumper''\footnote{\url{https://youtu.be/mw9cVvaM0vQ}}) starts with lifting his prosthetic leg; he jumps on this healthy leg for the whole episode using the prosthetic leg to keep balance. This is, definitely, not the most natural way of walking, but keeping balance with the prosthetic leg looks surprisingly natural.

\begin{figure}[t]
	\centering
	\includegraphics[width=\textwidth]{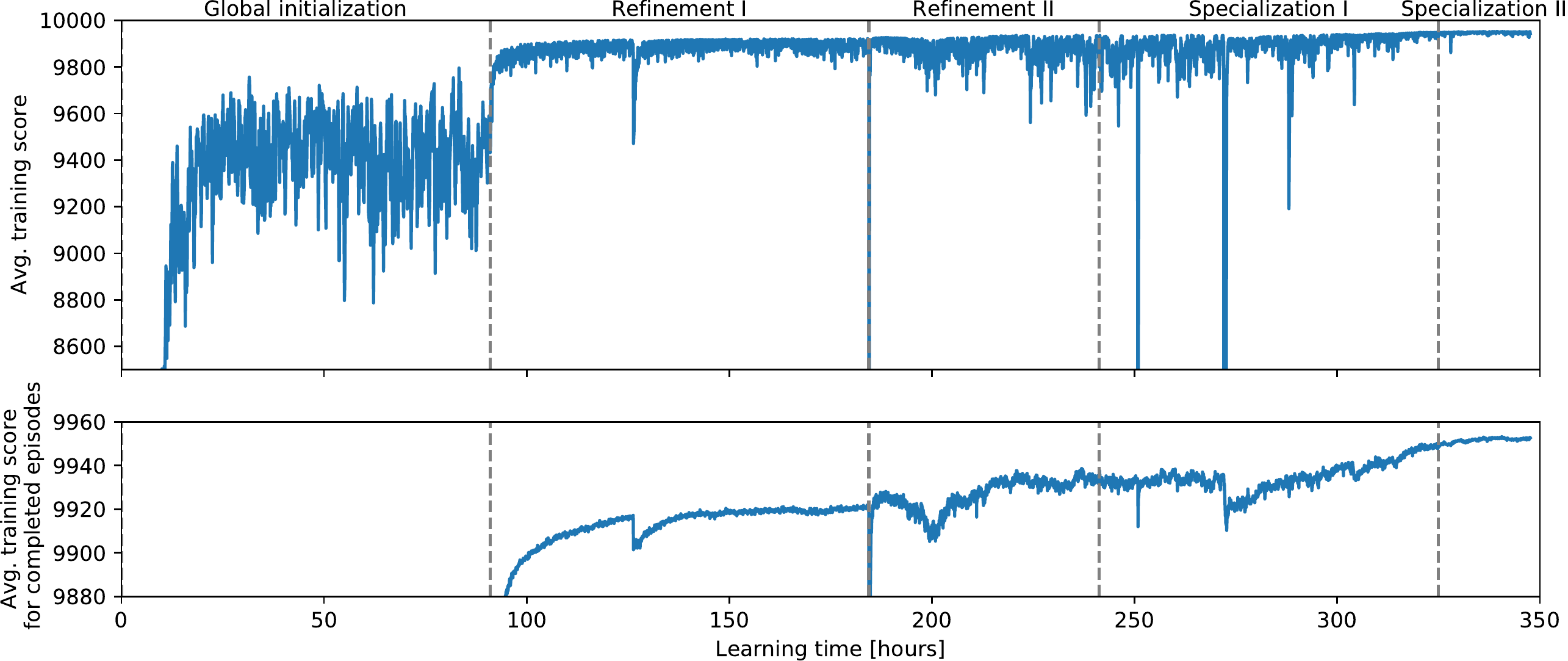}
	\caption{The training curve for the ``Dancer'' policy. The curve in the top figure is noisier since it shows the average score for all training episodes whereas the bottom one shows the score only for the completed episodes. Note that the performance during training is always worse than the performance during testing when the action noise is turned off.}
	\label{fig:nntraining}
\end{figure}

The training curve for the ``Dancer'' policy is shown in Fig.~\ref{fig:nntraining}.

\begin{table}[ht]
\centering
\caption{The performance of the trained policies}
\label{tab:nnresults}
\setlength\tabcolsep{0.15cm}
\setlength{\extrarowheight}{.1em}
\begin{tabular}{lr@{.}lr@{.}lr@{.}lr@{.}	lr@{.}l}
	\toprule
	\multirow{2}{*}{\bf Policy} & \multicolumn{4}{c}{\bf Score} & \multicolumn{6}{c}{\bf Avg. penalties}     \\
	\cmidrule(l){2-11}
	 & \multicolumn{2}{l}{\bf mean} & \multicolumn{2}{l}{\bf stdev} & \multicolumn{2}{l}{\bf velocity x} & \multicolumn{2}{l}{\bf velocity y} & \multicolumn{2}{l}{\bf activation} \\
	\cmidrule(l){1-11}
	Dancer  & 9954&5  & 6&7   & 30&9       & 7&6        & 6&9        \\
	Jumper  & 9949&0  & 15&1  & 32&3       & 12&9       & 5&8       \\
	\bottomrule
\end{tabular}
\end{table}


\subsection{Discussion}
The policy gradient-based method used by NNAISENSE to obtain efficient policies for this task is not particularly sample efficient. However, it requires only a little tuning and it robustly leads to well performing policies as evidenced in the ``Learning to Run'' 2017 and ``AI for Prostetics'' 2018 NeurIPS Challenges.

The peculiarity of the obtained gaits in this work is probably be due to the non-natural characteristic of the reward function used for this task. In the real world, humans are not rewarded for keeping a constant speed. Designing a reward function that leads to desired behaviour is known to be a challenging task in control and reinforcement learning.
\section{Distributed Quantile Ensemble Critic with Attention Actor}\label{s:dqec}
\sectionauthor{Sergey Kolesnikov, Oleksii Hrinchuk, Anton Pechenko}

Our method combines recent advances in off-policy deep reinforcement learning algorithms for continuous control, namely Twin Delayed Deep Deterministic Policy Gradient (TD3)~\cite{fujimoto2018addressing}, quantile value distribution approximation (QR-DQN)~\cite{dabney2017distributional}, distributed training framework~\cite{horgan2018distributed,barth2018distributed}, and parameter space noise with LayerNorm for exploration~\cite{plappert2017parameter}. We also introduce LAMA (last, average, max, attention~\cite{bahdanau2014neural}) pooling 
to 
take into account several temporal observations effectively. The resulting algorithm scored a mean reward of $9947.096$ in the final round and took \textbf{3rd place} in NeurIPS'18 AI for Prosthetics competition. We describe our approach in more detail below and then discuss contributions of its various components. Full source code is available at \url{https://github.com/scitator/neurips-18-prosthetics-challenge}.

\subsection{Methods}

\subsubsection{Twin Delayed Deep Deterministic Policy Gradient (TD3)}
TD3 algorithm is a recent improvement over DDPG which adopts Double Q-learning technique to alleviate overestimation bias in actor-critic methods. The differences between TD3 and DDPG are threefold. Firstly, TD3 uses a pair of critics which provides pessimistic estimates of Q-values in TD-targets (equation $10$ in~\cite{fujimoto2018addressing}).
Secondly, TD3 introduces a novel regularization strategy, target policy smoothing, which proposes to fit the value of a small area around the target action (equation $14$ in~\cite{fujimoto2018addressing}).
Thirdly, TD3 updates an actor network less frequently than a critic network (for example, one actor update for two critic updates). 

In our experiments, the application of the first two modifications led to much more stable and robust learning. Updating the actor less often did not result in better performance, thus, this modification was omitted in our final model.

\subsubsection{Quantile value distribution approximation}
Distributional perspective on reinforcement learning~\cite{bellemare2017distributional} advocates for learning the true return (reward-to-go) $Z_\theta$ distribution instead of learning a value function $Q_\theta$ only.
This approach outperforms traditional value fitting methods in a number of benchmarks with both discrete~\cite{bellemare2017distributional, dabney2017distributional} and continuous~\cite{barth2018distributed} action spaces.

To parametrize value distribution we use a quantile approach~\cite{dabney2017distributional} which learns $N$ variable locations and applies a probability mass of $\frac{1}{N}$ to each of them. The combination of quantile value distribution approximation and the TD3 algorithm is straightforward: first, we choose the critic network with minimum Q-value, and second, we use its value distribution to calculate the loss function and perform an update step:
\[
i^* = \arg\min_{i=1,2} Q_{\theta_i} (\mathbf{s}_{t+1},\mathbf{a}_{t+1}),\quad \mathcal{L}_{\theta_i} = \mathcal{L}^\text{quantile} \left( Z_{\theta_i}(\mathbf{s}_t,\mathbf{a}_t),r_t+\gamma Z_{\theta_{i^*}}(\mathbf{s}_{t+1},\mathbf{a}_{t+1}) \right),
\]
where $\mathbf{s}_t$ is the state and $\mathbf{a}_t$ is the action at the step $t$.

\subsubsection{Distributed training framework}\label{sss:distributed-jolly}
We propose the asynchronous distributed training framework\footnote{\url{https://github.com/scitator/catalyst}} which consists of training algorithms (trainers), agents interacting with the environment (samplers), and central parameter and data sharing server implemented as redis database. Unlike previous methods~\cite{horgan2018distributed,barth2018distributed} which use a single learning algorithm and many data collecting agents, we propose training several learning algorithms simultaneously with shared replay buffer. First of all, it leads to more diverse data, as a result of several conceptually different actors participating in a data collection process (for example, we can simultaneously train DDPG, TD3, and SAC~\cite{haarnoja2018soft}). Secondly, we can run several instances of the same algorithm with a different set of hyperparameters to accelerate hyperparameter selection process, which may be crucial in the case of limited resources.

\subsubsection{LAMA pooling}\label{sss:concatenation}
Sometimes the information from only one observation is insufficient to determine the best action in a particular situation (especially when dealing with partially observable MDP). Thus, it is common practice to combine several successive observations and declare a state using simple concatenation~\cite{mnih2015human} or more involved recurrent architecture~\cite{recurrent2018}. We introduce LAMA which stands for last-average-max-attention pooling -- an efficient way to combine several temporal observations into a single state with soft attention~\cite{bahdanau2014neural} in its core:
\[
\mathbf{H}_t = \{\mathbf{h}_{t-k},\dots,\mathbf{h}_t\},\quad \mathbf{h}^\text{lama}=[\mathbf{h}_t,\;\text{avgpool}(\mathbf{H}_t),\;\text{maxpool}(\mathbf{H}_t),\;\text{attnpool}(\mathbf{H}_t)].
\]

\subsubsection{Hybrid exploration}\label{sss:parameter-jolly}
We employ a hybrid exploration scheme which combines several heterogeneous types of exploration. 
With $70\%$ probability, we add random noise from $\mathcal{N}(0,\sigma I)$ to the action produced by the actor where $\sigma$ changes linearly from $0$ to $0.3$ for different sampler instances. With $20\%$ probability we apply parameter space noise~\cite{plappert2017parameter} with adaptive noise scaling, and we do not use any exploration otherwise. The decision of which exploration scheme to choose is made at the beginning of the episode and is not changed till its end.

\subsubsection{Observation and reward shaping}\label{sss:reward-jolly}
We have changed the initial frame of reference to be related to pelvis by subtracting its coordinates $(x,y,z)$ from all positional variables. In order to reduce inherent variance, we have standardized input observations with sample mean and variance of approximately $10^7$ samples collected during early stages of experiments. We have also rescaled the time step index into a real number from $[-1,1]$ and included it into observation vector as was recently proposed by~\cite{pardo2017time}.

At the early stages of the competition we have noticed that sometimes the learned agent tended to cross its legs as the simulator allowed one leg to pass through another. We have assumed that such behavior led to suboptimal policy and excluded it by introducing additional ``crossing legs'' penalty. 
Specifically, we have computed the scalar triple product of three vectors starting at pelvis and ending at head, left toe, and right prosthetic foot, respectively, which resulted in a penalty of the following form ($\vec{r}$ is a radius vector):
\[
p^\text{crossing legs} = 10 \cdot \min \left\{(\vec{r}^\text{head}-\vec{r}^\text{pelvis}, \vec{r}^\text{left}-\vec{r}^\text{pelvis}, \vec{r}^\text{right}-\vec{r}^\text{pelvis}), 0 \right\}.
\]
We have also rescaled the original reward with formula $r \leftarrow 0.1 \cdot (r - 8)$ to experience both positive and negative rewards, as without such transformation the agent always got positive reward (in the range of $\sim[7,10]$) which slowed learning significantly.

\subsubsection{Submit tricks}\label{sss:ensemble}

In order to construct the best agent from several learned actor-critic pairs we employ a number of tricks.

\textbf{Task-specific models}. Our experiments have revealed that most points are lost at the beginning of the episode (when agent needs to accelerate from zero speed to $1.25$ m/s) and when target velocity has big component on z-axis. Thus, we have trained two additional models for this specific tasks, namely ``start'' (which is active during the first $50$ steps of the episode) and ``side'' (which becomes active if the z-component of target velocity becomes larger than $1$ m/s).

\textbf{Checkpoints ensemble}. Adapting the ideas from~\cite{huang2017learning,dietterich2000ensemble,huang2017snapshot} and capitalizing on our distributed framework, we simultaneously train several instances of our algorithm with different sets of hyperparameters, and then pick the best checkpoints in accordance to validation runs on $64$ random seeds. Given an ensemble of actors and critics, each actor proposes the action which is then evaluated by all critics. After that, the action with the highest average Q-value is chosen.

\textbf{Action mixtures}. In order to extend our action search space, we also evaluate various linear combinations of the actions proposed by the actors. This trick slightly improves the resulting performance at no additional cost as all extra actions are evaluated together in a single forward pass.

\subsection{Experiments and results}
During our experiments, we have evaluated training performance of different models. For the complete list of hyperparameter and their values we refer the reader to our GitHub repository.

\begin{figure}[ht!]
\centering
    \begin{subfigure}[t]{0.49\textwidth}
    \centering
    \includegraphics[width=1\textwidth]{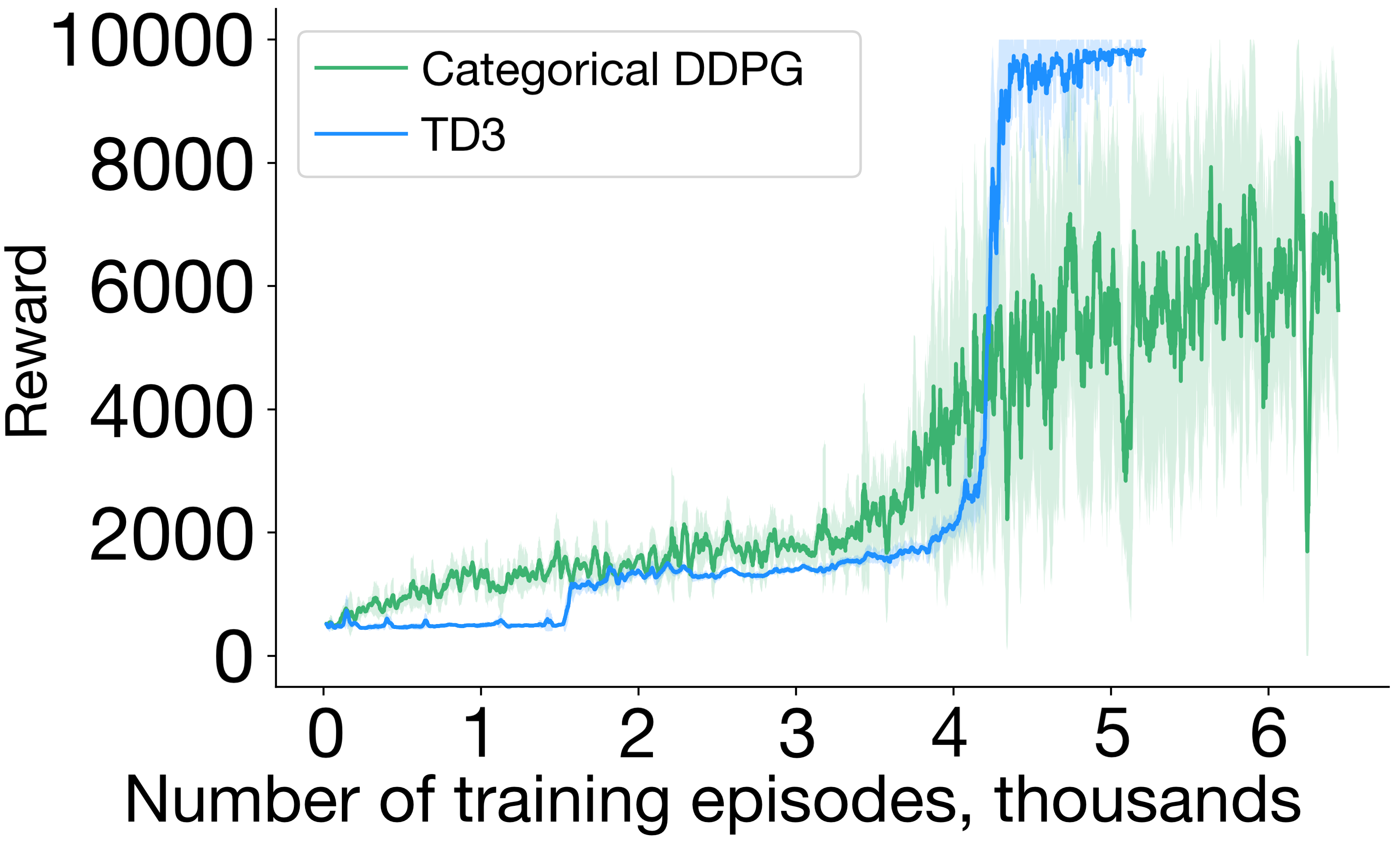}
    \caption{Comparison between TD3 and DDPG with categorical value distribution approximation (also known as D4PG~\cite{barth2018distributed}).}
    \label{fig:td3_vs_ddpg}
    \end{subfigure}\hfill
    \begin{subfigure}[t]{0.49\textwidth}
    \centering
    \includegraphics[width=1\textwidth]{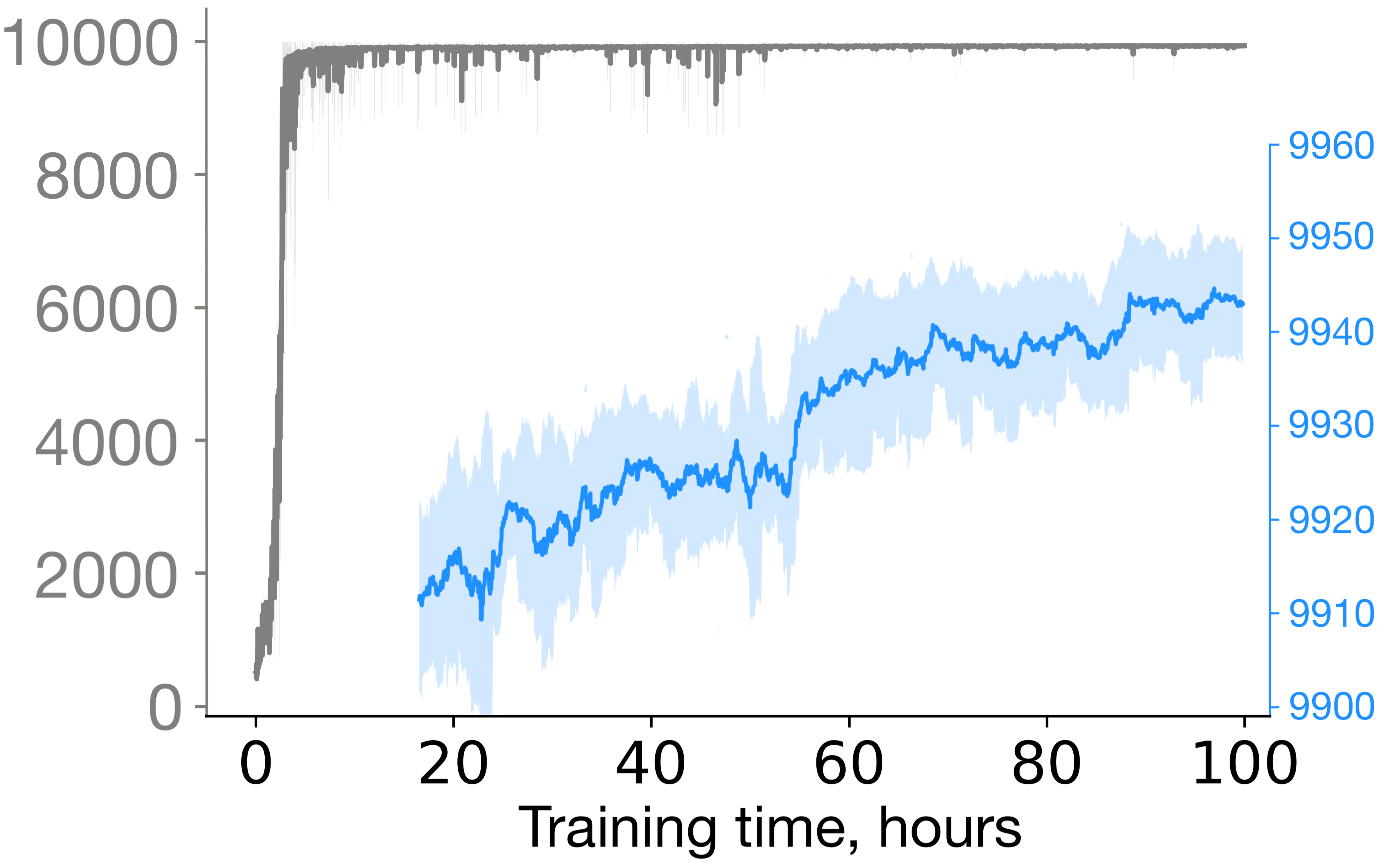}
    \caption{Learning curve for final model in the original scale (gray) and its rescaled version (blue).}
    \label{fig:jolly_roger_final}
    \end{subfigure}
    \caption{Learning curves for different approaches.}
    \label{fig:jolly_roger}
\end{figure}

\textbf{Model training performance}. Figure~\ref{fig:td3_vs_ddpg} shows a comparison of distributed TD3~\cite{fujimoto2018addressing} (without updating actor less often) and distributed DDPG with categorical value distribution approximation (also known as D4PG~\cite{barth2018distributed}). As we can see the TD3 exhibits much more stable performance which advocates for the use of two critics and fitting the value of a small area around the target action in continuous control. Figure~\ref{fig:jolly_roger_final} shows the learning curve for the final models we used in our solution. Although training to convergence takes quite a time ($4$ days), our method exhibits remarkable sample efficiency exceeding a score of $9900$ after just $10$ hours of training with $24$ parallel CPU samplers.

\begin{figure}[ht!]
\centering
    \begin{subfigure}[t]{0.49\textwidth}
    \centering
    \includegraphics[width=1\textwidth]{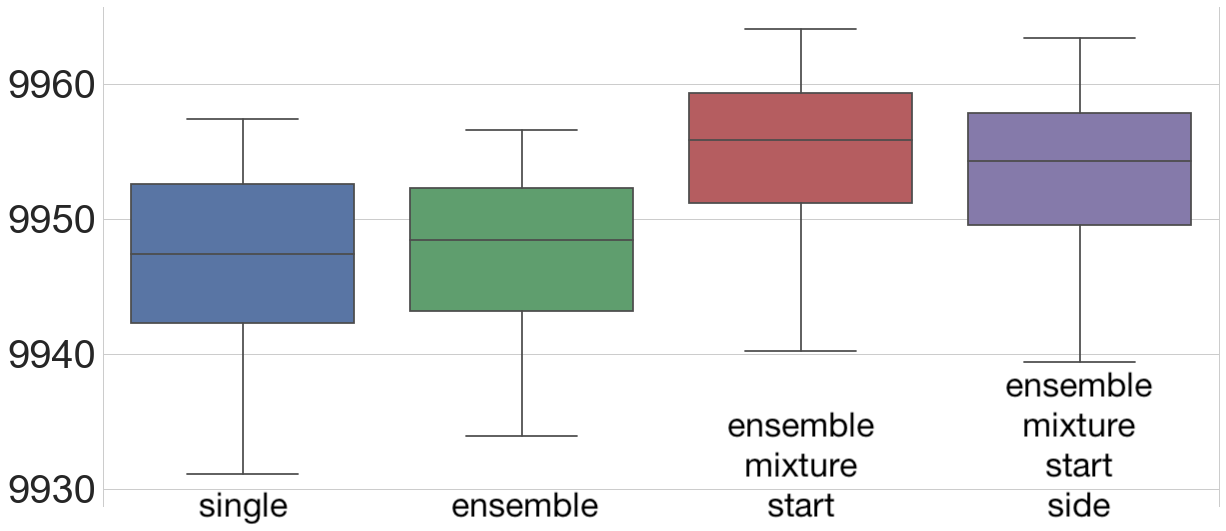}
    \caption{Performance on local evaluation, $64$ random seeds.}
    \label{fig:jolly_boxplot}
    \end{subfigure}\hfill
    \begin{subfigure}[t]{0.49\textwidth}
    \centering
    \includegraphics[width=1\textwidth]{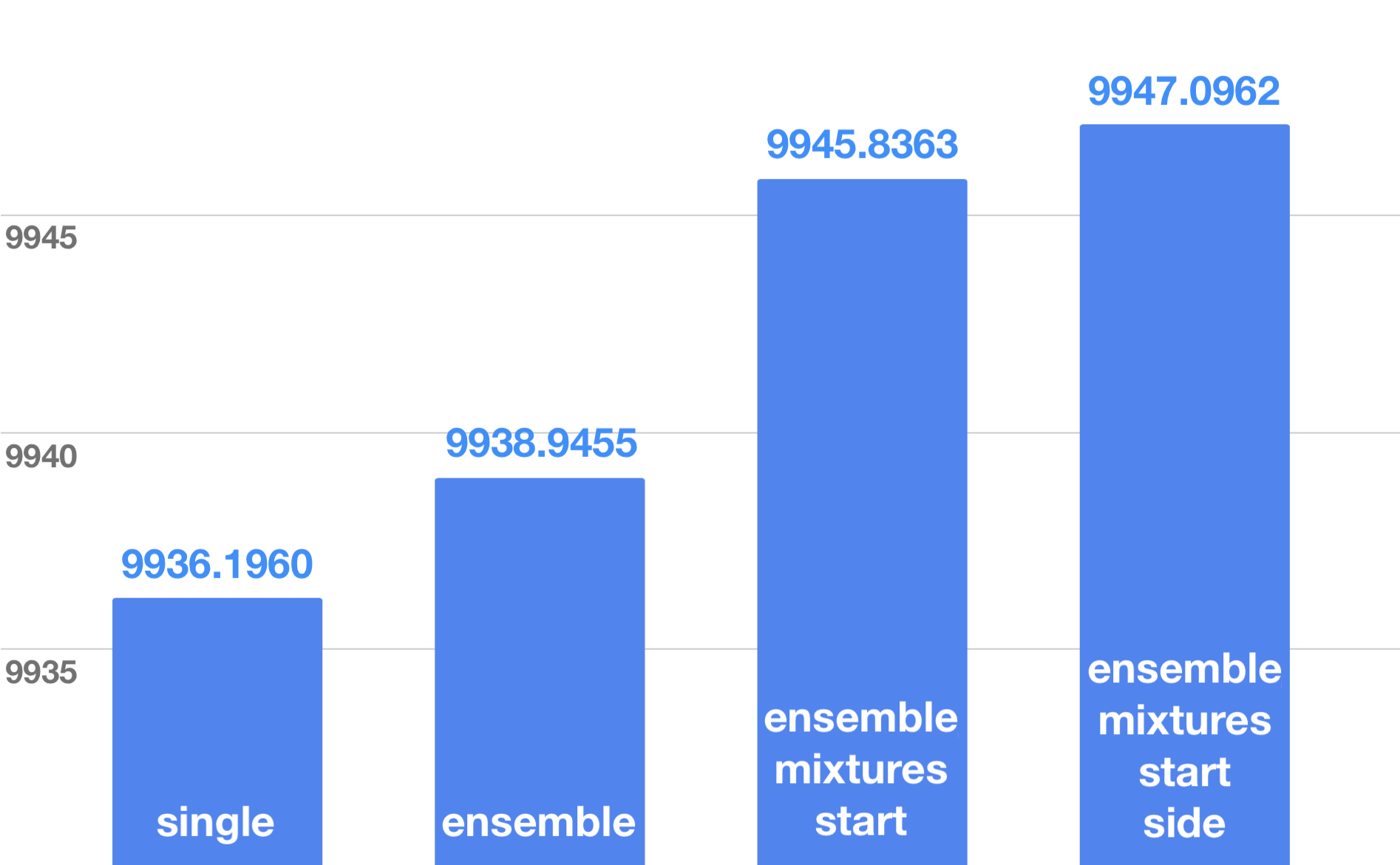}
    \caption{Performance on the final submit, $10$ seeds.}
    \label{fig:jolly_barplot}
    \end{subfigure}
    \caption{Performance of different submit tricks.}
    \label{fig:jolly_roger2}
\end{figure}

\textbf{Submit trick results}. Figure~\ref{fig:jolly_roger2} depicts the performance of our models with different submit tricks applied. The combination of all tricks allows us to squeeze an additional $11$ points from a single actor-critic model.

\subsection{Discussion}


We proposed the Distributed Quantile Ensemble Critic (DQEC), an off-policy RL algorithm for continuous control, which combines a number of recent advances in deep RL. Here we briefly summarize the key features of our algorithm and discuss the open questions from NeurIPS'18 AI for Prosthetics challenge.

\textbf{Key features.} Twin Delayed Deep Deterministic Policy Gradient (TD3)~\cite{fujimoto2018addressing}, quantile value distribution approximation~\cite{dabney2017distributional}, distributed framework~\cite{barth2018distributed} with an arbitrary number of trainers and samplers, LAMA (last, average, max, attention~\cite{bahdanau2014neural}) pooling, actor-critic ensemble~\cite{huang2017learning}.

\textbf{What could we do better?} First of all, we should analyze the particular features of the problem at hand instead of working on a more general and widely applicable approach. Specifically, we discovered that our agents fail on episodes with a high target velocity component on the z-axis only three days before the deadline and no time was left to retrain our models. If we found this earlier we could have updated our pipeline to train on less imbalanced data by repeating such episodes more often.

\textbf{What to do next?} Our model comprises a number of various building blocks. Although we consider all of them important for the final performance, a careful ablation study is required to evaluate contribution of each particular component. We leave this analysis for future work.


\section{Asynchronous DDPG with multiple actor-critics}\label{s:mattias}
\sectionauthor{Mattias Ljungström}

An asynchronous DDPG\citep{lillicrap2015continuous,silver2014deterministic} algorithm is setup with multiple actor-critic pairs. Each pair is trained with different discount factors on the same replay memory. Experience is collected asynchronously using each pair on a different thread. The goal of the setup is to balance time used on training versus simulation of the environment. The final agent scores 9938 on average over 60 test seeds, and placed fourth in the NeurIPS 2018 AI for Prosthetics competition.

\subsection{Methods}

\subsubsection{Multiple Actor-Critic pairs}

Simulation of the given environment is extremely costly in terms of CPU. Each step can require between 0.1s to 20 minutes to complete. To optimize utilization of CPU during training, an asynchronous approach is beneficial. During round 1 of the competition, a DDPG algorithm was used in combination with asynchronous experience collection on 16 threads. Training was done on a physical computer with 16 cores. Analysis of CPU utilization during training showed that only a fraction of CPU was used to train the network, and most time (>95\%) was spent on environment simulation.

To shift this more in favor of network training, a multiple of actor-critic (AC) pairs were trained on the same experience collected. Each AC pair has a different discount factor, and is trained with unique mini-batches. After a set number of steps, actors are shifted in a circular way to the next critic. During training each AC takes turns to run episodes on 16 threads. To support this larger network setup, training is done on GPU instead of CPU. All experience is stored into the same replay memory buffer. For the final model, 8 pairs of ACs were used, to balance CPU but also due to GPU memory limits.

\begin{figure}[ht!]
    \centering
    \begin{subfigure}[t]{0.6\textwidth}
        \centering
        \includegraphics[width=1\textwidth]{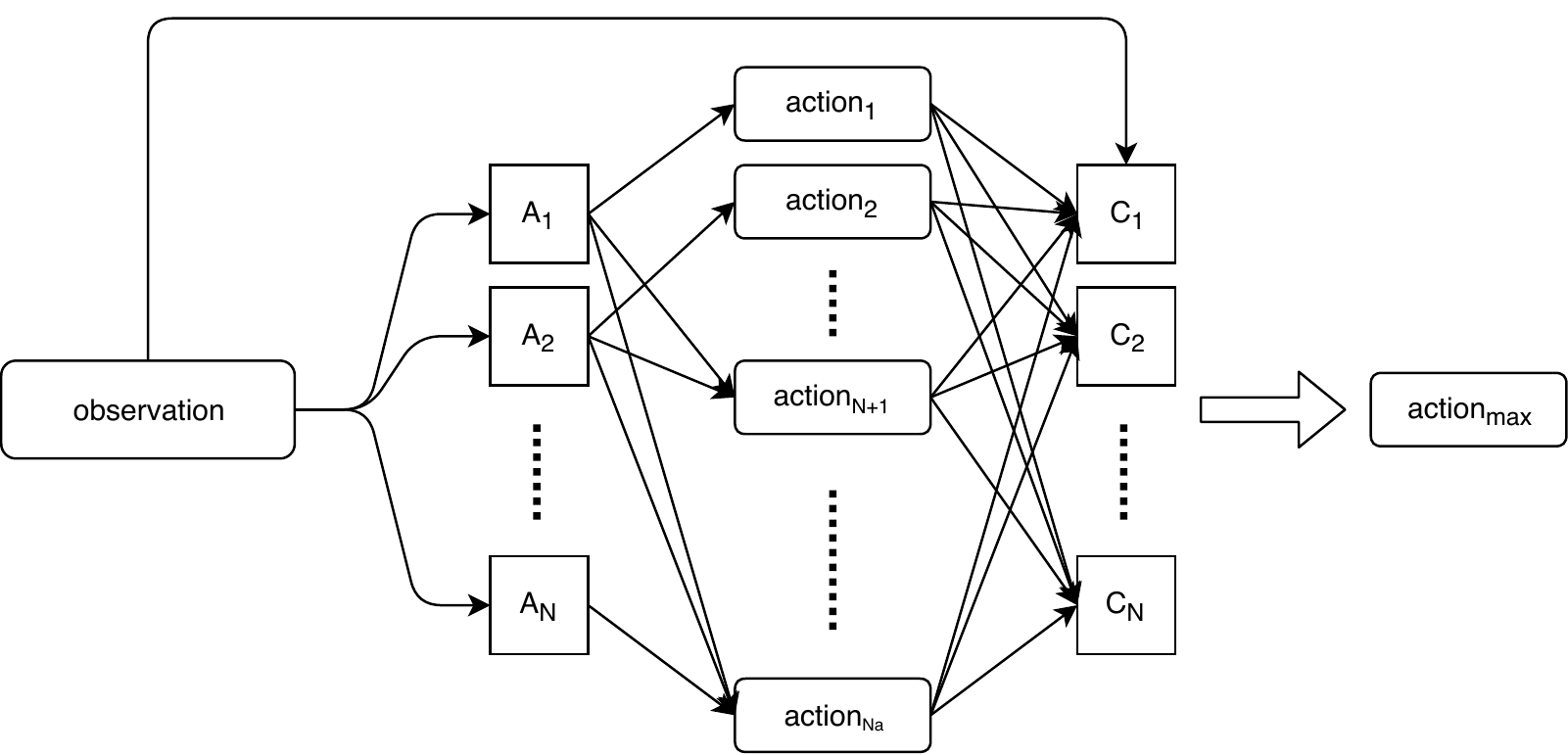}
        \caption{Inference with Actor-Critic pairs}
        \label{img:ac-pairs}
    \end{subfigure}\hfill
    \begin{subfigure}[t]{0.35\textwidth}
        \centering
        \includegraphics[width=1\textwidth]{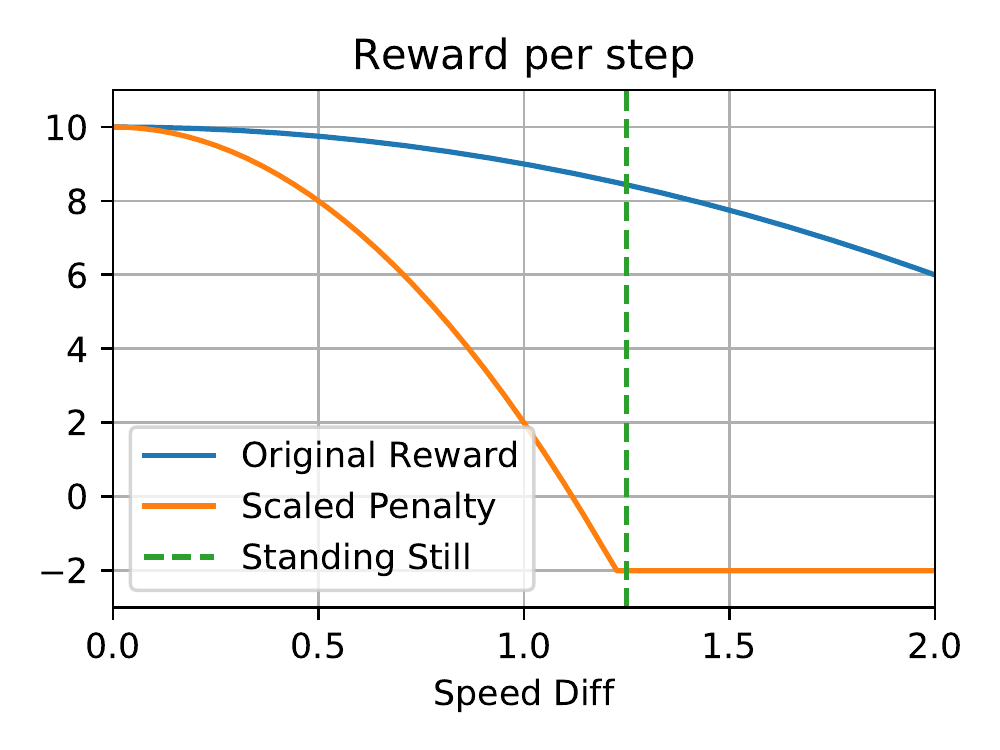}
        \caption{Effect of penalty scaling}
        \label{img:penalty-scaling}
    \end{subfigure}
    \caption{Actor-Critic pairs and reward penalties}
    \label{fig:setup-penalty}
\end{figure}

At inference, each actor produces an action based on the current observation. Further new actions are created from averages of 2 and 3 of the initial actions, and are added to a list of potential actions. Potential actions are evaluated by all critics, and the action with the maximum value is picked (Fig.~\ref{img:ac-pairs}).

Complete source code for solution is available at \url{https://github.com/mattiasljungstrom/learningtorun_2018}.

\subsubsection{Reward Shaping}\label{sss:reward-mattias}

The original reward has a very low penalty for not moving. This resulted in agents that were satisfied with standing still. To counter this the original penalty was multiplied with 8, but capped to a maximum of 12. The purpose was to give standing still a slightly negative reward. Moving at the correct velocity would still reward a value of 10. See Fig.~\ref{img:penalty-scaling}.

Furthermore, additional penalties were added to the reward. A penalty for not bending the knees turned out to be helpful in making the agent learn to walk in a more natural way. A penalty was added to ensure that each foot was below the upper part of the respective leg of the skeleton. This helped the agent avoid states were legs were pointing sideways. The pelvis orientation was penalized for not pointing in the velocity direction. This helped the agent to turn correctly. To further encourage correct turns, a penalty for not keeping the feet in the target velocity direction was added. Finally, a penalty was added to avoid crossing legs and feet positions, as this usually meant the skeleton would trip over itself.

Only penalties were used, since trials with positive rewards showed that the agent would optimize towards fake rewards. With pure penalties, the agent is only rewarded for actions that would also lead to good scores in a penalty free setup. The total sum of penalties was capped at -9 to keep the final reward in the range of [-11, 10].

\subsubsection{Replay Memory Optimization}

During initial tests it became clear that a too small replay memory would lead to deteriorating performance of the agent when early experience was overwritten. To counter this, a very large memory buffer was used. Development time was spent optimizing the performance of this memory buffer, and changing from a dynamic buffer to a static buffer increased training performance 10 times for large buffer sizes. The size of the buffer used in final training was 4 million experiences. 

\subsubsection{Hyper parameters, Environment Changes and Observation Shaping}\label{sss:observation-mattias}

Due to limited compute resources, a full hyperparameter search was not feasible. A few parameters were evaluated during training for 24 hours each. Discount factors between 0.95 and 0.99 were evaluated. Trials showed the agent learned to walk faster using a value range of [0.96, 0.976]. Different rates of learning rate per step were evaluated. Evaluating mini-batch sizes from 32 to 128 showed that a higher value was more beneficial.

During training the environment was tweaked so that the agent would be forced to turn 6 times instead of 3. The agent would also skip every other step during training, so that experience was collected at 50Hz instead of 100Hz.

All observation values were manually re-scaled to be in an approximate range of [-5, 5]. All positions, rotations, velocities and accelerations were made to be relative towards pelvis. The target velocity was reshaped into a speed and relative direction change.

\subsection{Experiments and results}
\subsubsection{Refinement Training}

After training the final model for 22000 initial episodes, the average score during testing was 9888. At this point a series of evaluations were done on 60 test seeds. It was discovered that the agent would still fall in certain situations.

\begin{figure}[ht!]
    \centering
    \begin{subfigure}[t]{0.45\textwidth}
        \centering
        \includegraphics[width=1\textwidth]{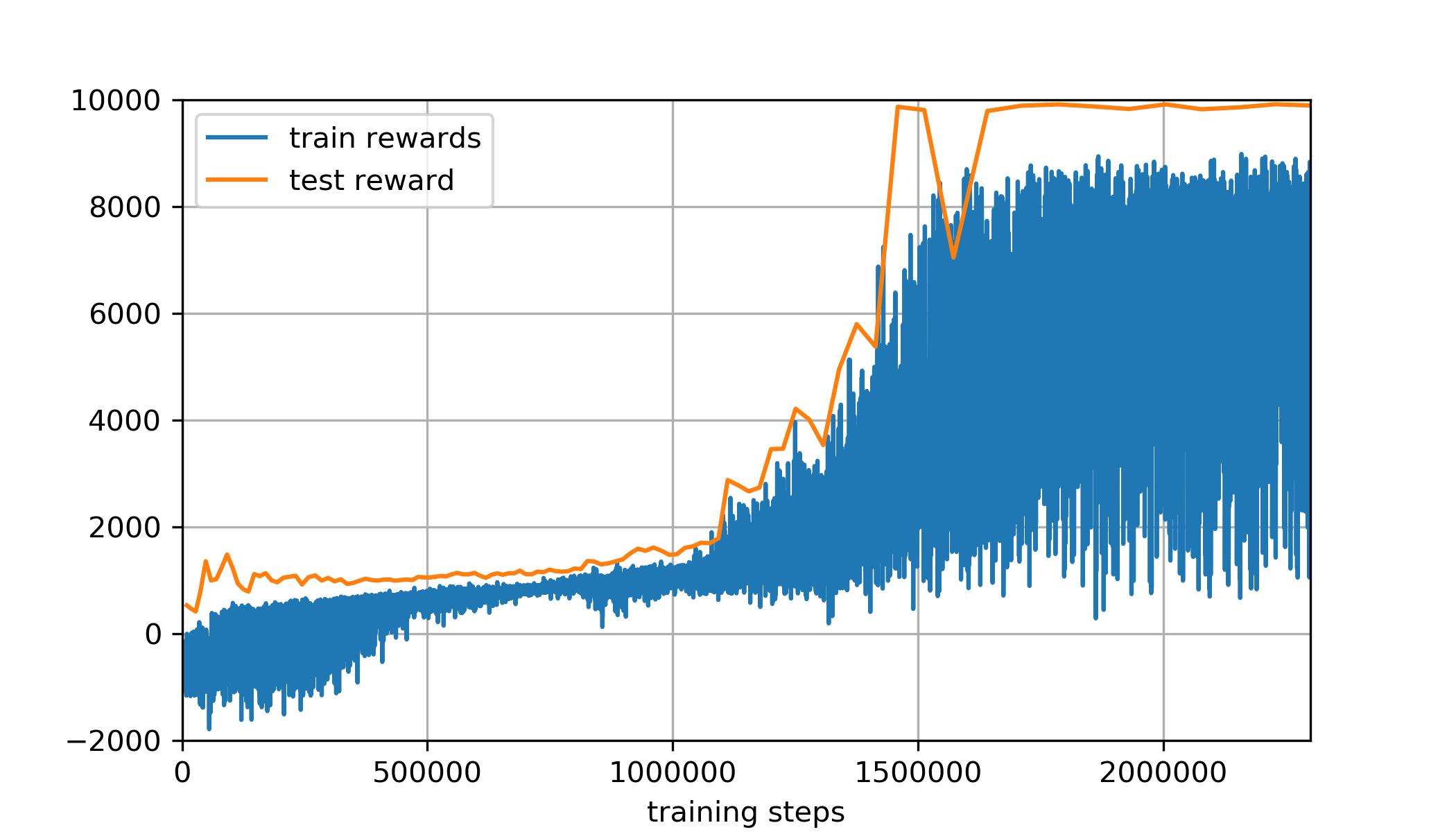}
        \caption{Rewards during initial 22000 episodes}
        \label{fig:training-rewards}
    \end{subfigure}
    \begin{subfigure}[t]{0.47\textwidth}
        \includegraphics[width=1\textwidth]{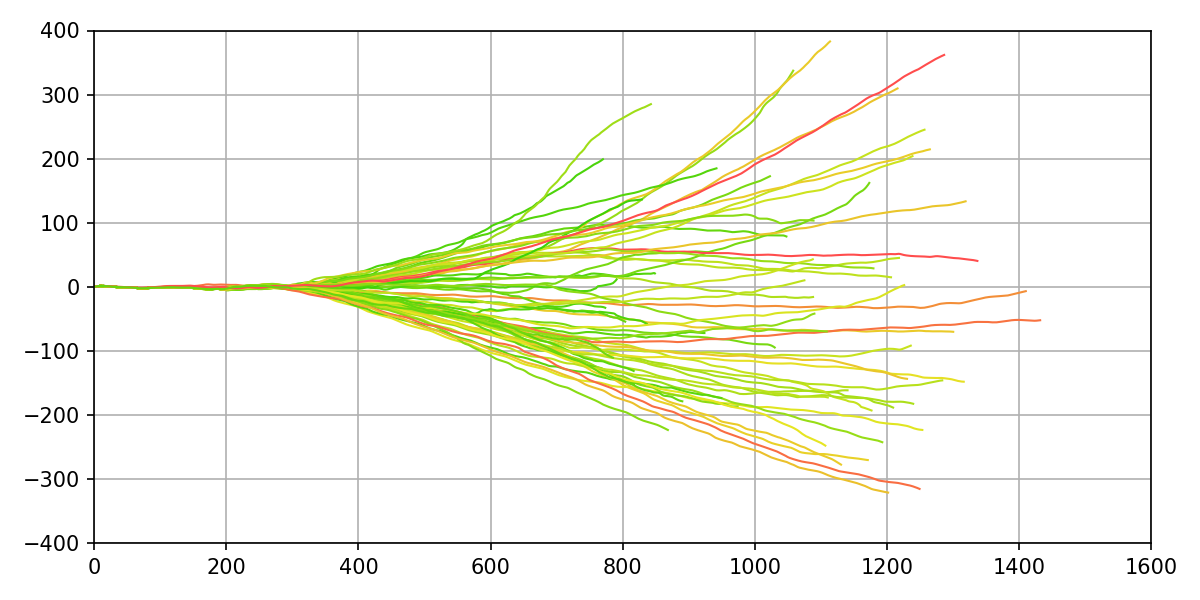}
        \caption{Path plot of initial model tested on 60 seeds. Color [red, green] maps to rewards [9800, 9950]}
        \label{fig:test-initial-60}
    \end{subfigure}
    \caption{Training statistics}
    \label{fig:training}
\end{figure}

From these 60 tests the worst performing seeds were selected. The agent was run on these to collect experience without training for a thousand episodes. Training was then continued using only seeds with low scores. In addition, learning rates for both actor and critic were lowered to allow fine-tuning. After another 9000 episodes of training in this way, the average score of the agent was 9930.

At this point training was switched from turning 6 times per episode to 3 times, as in the final evaluation. Again, new experience was collected without training, and training was then continued using 3 turns. After approximately 4000 more training episodes, the final agent\footnote{\url{https://mljx.io/x/neurips_walk_2018.gif}} scored 9938.7 on average over 60 test seeds.

\subsection{Discussion}
Using multiple actor-critic pairs allow the agent to learn more efficiently on the experience collected. Refinement training with human-assisted adjustments enabled agent to go from average score of 9888 to 9938. Reward penalties allow the agent to learn to walk faster by excluding known bad states, but probably limited exploration which could have generated better rewards.

\section{ApeX-DDPG with Reward Shaping and Parameter Space Noise}\label{s:apexddpg}
\sectionauthor{Zhen Wang, Xu Hu, Zehong Hu, Minghui Qiu, Jun Huang}

We leverage ApeX~\citep{horgan2018distributed}, an actor-critic architecture, to increase the throughput of sample generation and thus accelerate the convergence of the DDPG~\citep{lillicrap2015continuous,silver2014deterministic} algorithm with respect to wall clock time.
In this way, a competitive policy, which achieved a $9900.547$ mean reward in the final round, can be learned within three days.
We released our implementation\footnote{\url{https://github.com/joneswong/rl_stadium}} which reuses some modules from Ray~\citep{moritz2018ray}.
Based on a standard ApeX-DDPG, we exploited reward shaping and parameter space noise in our solution, both of which bring in remarkable improvements.
We will describe these tricks thoroughly in this section.

\subsection{Methods}
\subsubsection{ApeX}\label{sss:distributed-apex}
\begin{figure}[ht!]
\centering
    \begin{subfigure}[t]{0.49\textwidth}
    \centering
    \includegraphics[width=1\textwidth]{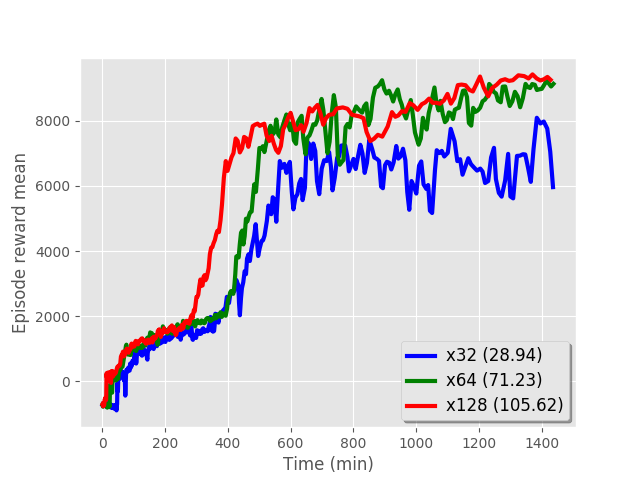}
    \caption{Comparison of different \#actor}
    \label{fig:cmpnumactor}
    \end{subfigure}\hfill
    \begin{subfigure}[t]{0.49\textwidth}
    \centering
    \includegraphics[width=1\textwidth]{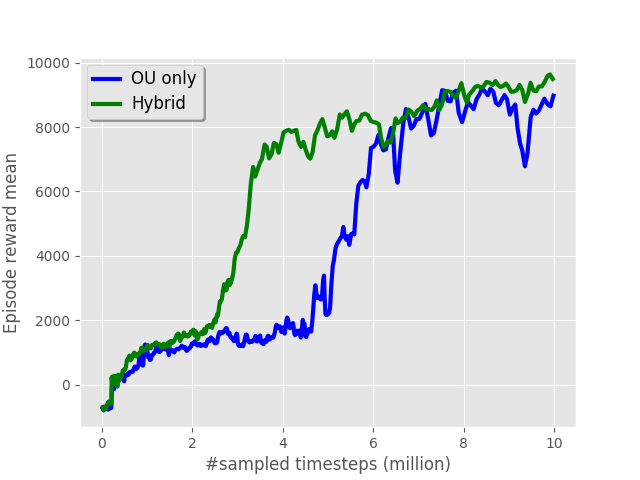}
    \caption{Comparison of exploration strategies}
    \label{fig:cmpexplorations}
    \end{subfigure}
    \label{fig:cmp}
    \caption{Effectiveness of ApeX and parameter space noise}
\end{figure}

In the ApeX architecture, each actor interacts with its corresponding environment(s) and, once a batch of samples has been generated, sends the samples to the learner.
Meanwhile, each actor periodically pulls the latest model parameters from learner.
The learner maintains collected samples in a prioritized replay buffer~\citep{schaul2015prioritized} and continuously updates model parameters based on mini-batches sampled from the buffer.
osim-rl~\citep{kidzinski2018learningtorun} consumes around $0.22s$ for simulating one step where the actor side is the bottleneck and the throughput (time steps per second) as well as convergence speed increases significantly as we add more actors (see Fig.~\ref{fig:cmpnumactor}).

\subsubsection{Reward Shaping}\label{sss:reward-itshigh}
\begin{figure}[ht!]%
        \centering
        \begin{subfigure}[t]{0.23\textwidth}
        \centering
                \includegraphics[width=1\textwidth]{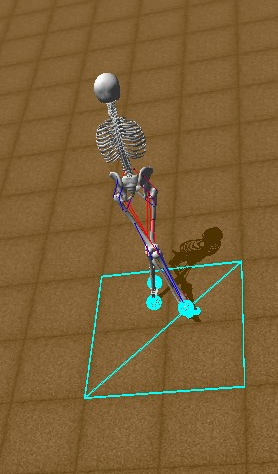}
                \caption{Without Penalty}
                \label{fig:scissors}
        \end{subfigure}\hfill
        \begin{subfigure}[t]{0.45\textwidth}
        \centering
                \includegraphics[width=1\textwidth]{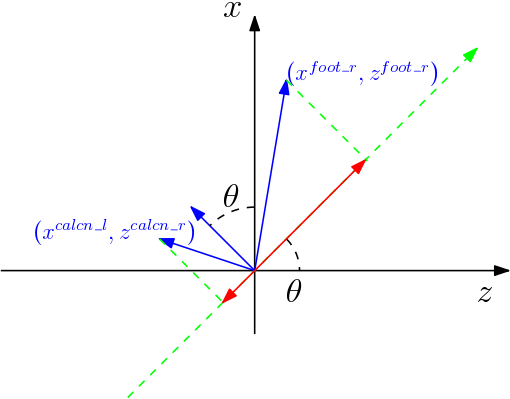}
                \caption{Geometric Intuition}
                \label{fig:pgeometric}
        \end{subfigure}\hfill
        \begin{subfigure}[t]{0.195\textwidth}
        \centering
                \includegraphics[width=1\textwidth]{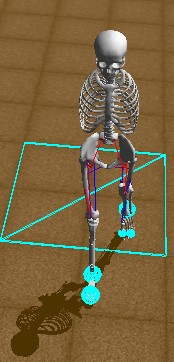}
                \caption{With Penalty}
                \label{fig:counterscissors}
        \end{subfigure}
        \caption{Eliminate scissor leg via reward shaping}
\end{figure}

osim-rl favors agents that walk at a target velocity, regardless if the gait is unnatural.
Nonetheless, we assume that the optimal policy does not use an unnatural gait and thus, in order to trim the search space, we shape the original reward for encouraging our agent to walk in a natural way.
First, we noticed that our agent is inclined to walk with scissor legs (see Fig.~\ref{fig:scissors}).
With this gait, agents become extremely brittle when the target orientation substantially changed.
We remedy scissor legs by adding a penalty to the original reward:
\begin{equation}
p^{\text{scissors}}=[x^{\text{calcn\_l}}\sin\theta+z^{calcn\_l}\cos\theta]_{+}+[-(x^{\text{foot\_r}}\sin\theta+z^{foot\_r}\cos\theta)]_{+}
\end{equation}
where $\theta$ is the rotational position of pelvis about the $y$-axis, $[x]_{+}\triangleq\max(x,0)$, and all positions are measured in a relative way with respect to pelvis positions.
We show the geometric intuition of this penalty in Fig.~\ref{fig:pgeometric}.
Some case studies confirmed its effectiveness (see Fig.~\ref{fig:counterscissors}).
Another important observation is that, at the early stage of a training procedure, the subject follows the target velocity by walking sideways, leading to a residual between the current heading and the target direction.
This residual may accumulate for each time the target velocity changes, e.g., the subject persists heading in $x$-axis and consecutively encounters changes that all introduce increments in the positive $z$-axis.
Intuitively speaking, heading changes that exceed the upper bound of osim-rl (i.e., $\frac{\pi}{8}$) are intractable.
Thus, we define a penalty as below to avoid a crab walk:
\begin{equation}
p^{\text{sideways}}=1-\frac{(v_{x}\cos\theta-v_{z}\sin\theta)}{\sqrt{v_{x}^{2}+v_{z}^{2}}}
\label{eq:sideways}
\end{equation}
where $v_x,v_z$ stand for the target velocity in $x$ and $z$ axes respectively.
The RHS of \eqref{eq:sideways} is the cosine distance between the target velocity and the pelvis orientation in the $x,z$-plane.
Our solution got $9828.722$ reward at the most difficult episode of the final round (i.e., episode3) which potentially consists of large heading changes.
To the best of our knowledge, our gap between this reward and the mean reward is smaller than that of many other competitors which strongly indicates the usefulness of this penalty.

\subsubsection{Parameter Space Noise}\label{sss:parameter-itshigh}
In our solution, the actors explore by either posing Ornstein--Uhlenbeck (OU) noise~\citep{lillicrap2015continuous} upon the actions or adding a Gaussian perturbation to the model parameters~\citep{plappert2017parameter}, with a fifty-fifty chance~\citep{pavlov}.
For each actor, the OU samples are multiplied by an actor-specific coefficient in analogy to taking $\epsilon$-greedy exploration with various $\epsilon$ at the same time.
On the other hand, the parameter space noise enriches the behavior policies, without which the policies of different actors are close to each other at each certain moment.
We present the advantages of such a hybrid exploration strategy in Fig.~\ref{fig:cmpexplorations}.

\subsection{Experiments and results}
We trained our model on the Alibaba Cloud PAI platform~\footnote{\url{https://www.alibabacloud.com/press-room/alibaba-cloud-announces-machine-learning-platform-pai}}.
For ApeX, we used 1 learner (x1 P100 GPU, x6 CPU) and 128 actors (x1 CPU).
For DDPG, our configuration can be found at \textit{/examples/r2\_v4dot1b.json} in our repository.
There are mainly two peculiarities that need to be clarified.
First, distinguishing from the critic network used in original DDPG, we apply a fully-connected layer (128 neurons) to the action before concatenating the action and state channels.
We argue that, in high-dimensional continuous control like osim-rl, the action also needs such a feature extraction procedure for better approximating the Q function.
Empirical evaluation confirmed our point.
Second, we noted that there are three target velocity changes within 1000 time steps on average.
Time steps when such changes occur are accidental observations to the agent.
Such "noisy" samples are likely to import oscillations which can often be alleviated by increasing the batch size.
We made rough comparisons among batch sizes of $\{256, 512, 1024\}$ and the results support using the largest one.

\subsection{Discussion}
In addition to the posture features, we lumped a normalized time step into our state representations.
At the first glance, this seems redundant, as the principles of running doesn't change whichever step the subject is at.
However, the time step feature advanced the mean reward of our solution from around $9800$ to $9900$.
We regard its contribution as variance reduction of V/Q-values.
Without this feature, the V/Q-values of the same state (i.e., posture) decrease along time step, since this challenge considers a finite horizon.
We think distributional Q-learning is an alternative, and how to combine it with deterministic policy gradient deserves further investigation.
\section{Deep Reinforcement Learning with GPU-CPU Multiprocessing}\label{s:jbr}
\sectionauthor{Aleksei Shpilman, Ivan Sosin, Oleg Svidchenko, Aleksandra Malysheva, Daniel Kudenko}

One of the main challenges we faced is that running the simulation is very CPU-heavy, while the optimal computing device for training neural networks is a GPU. One way to overcome this problem is building a custom machine with GPU to CPU proportions that avoid bottlenecking one or the other. Another is to have the GPU machine (such as AWS accelerated computing instance) work together with the CPU machine (such as the AWS compute optimized instance). We have designed a framework for such a tandem interaction \cite{ivan_sosin_2018_1938263}.

For the AI in Prosthetics competition we used the DDPG algorithm \cite{lillicrap2015continuous} with 4 layers of 512 neurons in the actor network and 4 layers of 1024 neurons in the critic network. We also performed additional feature engineering, two-stage reward shaping, and ensembling through SGDR \cite{loshchilov2017SGDR}. 

\subsection{Methods}

\subsubsection{GPU-CPU Multiprocessing}\label{sss:distributed-jbr}

\begin{figure}
\caption{Framework for running processes on a tandem GPU (client) and CPU (server) machines.}
\label{img:jbr-framework}
    \begin{center}
        \includegraphics[width=0.7\textwidth]{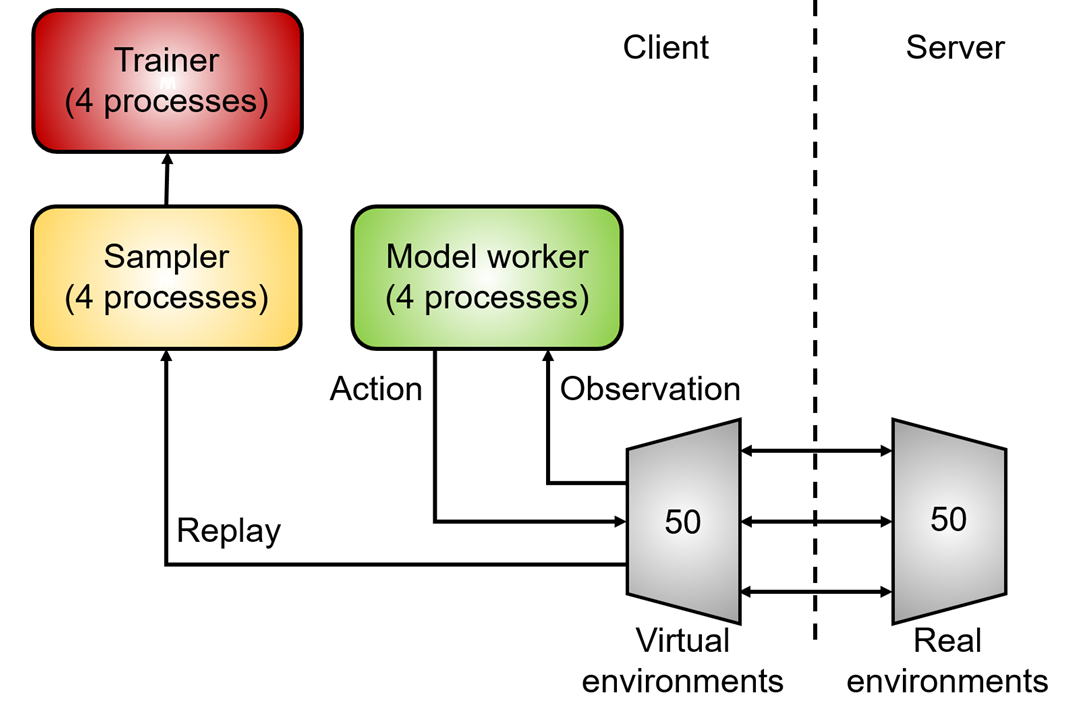}
    \end{center}
\end{figure}

Figure \ref{img:jbr-framework} shows our training framework. We divide it into a client and a server side. The client (GPU instance) trains the model based on data received from the server (CPU instance). On the server side we launch a number of real environments wrapped in a HTTP-server to run the physical simulation. On the client side we launch a corresponding number of virtual environments that redirect requests to OpenSim environments. These virtual environments transmit the state (in queue) to model workers that process the state and output the actions. Model workers' networks are constantly updated by the trainer via shared memory.
Samplers handle complete episodes and produce a batch for trainers to train the actor and critic networks on.

\subsubsection{Additional tricks}\label{sss:observation-jbr}

We used the DDPG algorithm \cite{lillicrap2015continuous} with the following methodologies that seem to improve the final result:

\textbf{Feature engineering}. In addition to the default features, we have engineered the following additional features:
\begin{itemize}
    \item XYZ coordinates, velocity, and acceleration relevant to the pelvis center point, body point, and head point ($10\times3\times3\times3=270$ features).   
    \item XYZ rotations, rotational velocities and rotational accelerations relevant to the pelvis center point ($10\times3\times3=90$ features).
    \item XYZ coordinates, velocities, and accelerations of center of mass relevant to the pelvis center point ($3\times3=9$ features).
\end{itemize}

\begin{figure}
\caption{Improvement of training speed and performance after engineering additional features.}
\label{img:jbr-fe}
    \begin{center}
        \includegraphics[width=0.7\textwidth]{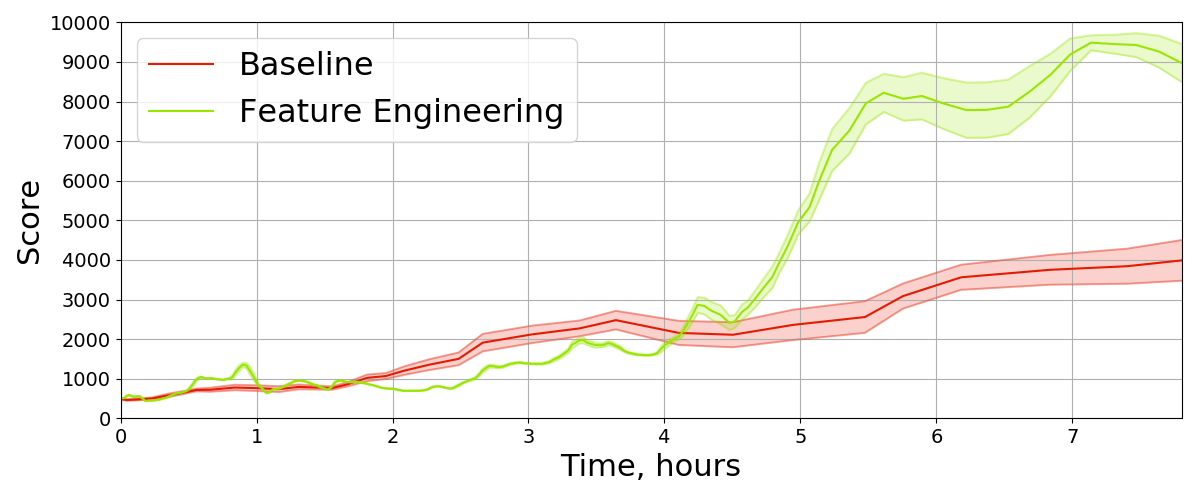}
    \end{center}
\end{figure}

The size of the resulting feature vector was 510. Figure \ref{img:jbr-fe} shows the result of adding reward shaping and new features to the baseline DDPG algorithm.

\textbf{Reward shaping}. We used two-stage reward shaping. For the first 7 hours we used the following reward function, that is much easier to train with because it is less punishing at the beginning of training, than the original Round 2 reward function

\begin{equation}
r = 1 - \frac{||v_{target} - v|| ^ 2}{||v_{target}||^2},
\end{equation}

where $v_{target}$ and $v$ is the target velocity and actual velocity, respectively. After that we used a modified and clipped Round 2 reward function: 

\begin{equation}
    r = 
\begin{cases} 
2 \cdot r_{origin} - 19&\text{if }\ r_{origin} \in (9.5, 10)\\
-1 & \text{otherwise,}
\end{cases}
\end{equation}

where $r_{origin}$ is the original Round 2 reward function. This reward function awards the model for a high score and penalizes any score below 9.5. 

\textbf{Ensembles}. We used Stochastic Gradient Descent with Warm Restarts (SGDR \cite{loshchilov2017SGDR}) to produce an ensemble of 10 networks, and then we chose the best combination of 4 networks by grid-search. The final action was calculated as an average output vector of those 4 networks.

\subsection{Experiments and results}

We used an AWS p3.2xlarge instance with Nvidia Tesla V100 GPU and 8 Intel Xeon CPU cores for training the network in tandem with a c5.18xlarge instance with 72 CPU cores (50 were utilized) for running the simulations. We used 50 environments (real and virtual), 4 Model workers, Samplers and Trainers. We trained models for ~20 hours with an additional ~10 hours used for SGDR ensembling.

The results presented in Table \ref{tbl:jbr-tricks} show confidence intervals for the score, calculated on 100 different random seeds for the Round 2 reward. The final score that was achieved on the ten seeds used by the organizers was 9865.  

\begin{table}
\centering
\begin{tabular}{|c|c|c|}
\hline
Model & Score & Frequency of falling \\
\hline
Baseline & 4041.10 $\pm$ 539.23 & N/A \\
\hline
Feature Engineering & 8354.70 $\pm$ 439.3 & 0.36 \\
\hline
Feature engineering + Reward Shaping & 9097.65 $\pm$ 344 & 0.21 \\
\hline
Feature engineering + Reward Shaping + Ensembles & 9846.72 $\pm$ 29.6 & 0.02 \\
\hline
\end{tabular}
\caption{Performance of models with different modules on 100 random seeds for Round 2 rewards}
\label{tbl:jbr-tricks}
\end{table}

\subsection{Discussion}

Our framework allows for the utilization of GPU and CPU machines in tandem for training a neural network. Using this approach we have been able to train an ensemble of networks that achieved the score of 9865 (6th place) in only 20 hours (+10 hours for ensembling with SGDR) on tandem p3.2xlarge-c5.18xlarge instances.

Our code is available at \url{https://github.com/iasawseen/MultiServerRL}.

\section{Model-guided PPO for sample-efficient motor learning}\label{s:lance}
\sectionauthor{Lance Rane}

Proximal policy optimisation (PPO) \cite{schulman2017proximal} has become the preferred reinforcement learning algorithm for many due to its stability and ease of tuning, but it can be slow relative to off-policy algorithms. Leveraging a model of the system’s dynamics can bring significant improvements in sample efficiency. 

We used human motion data in combination with inverse dynamics and neuromusculoskeletal modelling to derive, at low computational cost, guiding state trajectories for learning. The resulting policies, trained using PPO, were capable of producing flexible, lifelike behaviours with fewer than 3 million samples.

\subsection{Methods}
We describe methods and results for the final round of the competition only, where the task was to train an agent to match a dynamically changing velocity vector.

\subsubsection{State pre-processing and policy structure}
Positional variables were re-defined relative to a pelvic-centric coordinate system, to induce invariance to the absolute position and heading of the agent. The state was augmented with segment and centre of mass positions, velocities and accelerations, muscle fibre lengths and activation levels, and ground contact forces prior to input to the policy network. \textit{x} and \textit{z} components of the target velocity were provided in both unprocessed form and also after subtracting corresponding components of the current translational velocity of the agent. All values were normalized using running values for means and standard deviations.

The policy was a feedforward neural network with 2 layers of 312 neurons each, with \textit{tanh} activation functions. The action space was discretized such that the output of the network comprised a multicategorical probability distribution over excitations for each of the 18 muscles.

\subsubsection{Phase 1: Policy initialisation}

During phase 1, training was supported by the use of guiding state trajectories. Motion data describing multiple gait cycles of human non-amputee straight-line walking at 1.36m/s \cite{john2013stabilisation} were processed by resampling to generate three distinct sets of marker trajectories corresponding to average pelvic translational velocities of 0.75, 1.25 and 1.75m/s. For each of these three datasets the following pipeline was executed:

\begin{itemize}
	\item Virtual markers corresponding to the experimental markers were assigned to the OpenSim prosthetic model, except those for which no parent anatomy existed in the model.   
	\item Inverse kinematic analysis was performed to compute a trajectory of poses (and resultant joint angles) of the model that closely matched the experimental data. 
	\item Computed muscle control (CMC) \cite{thelen2003generating} was used to derive trajectories of states and muscle excitations consistent with the motion. 
\end{itemize}

CMC performs inverse dynamic analysis followed by static optimization with feedforward and feedback control to drive a model towards experimental kinematics. As a method for finding controls consistent with a given motion, it may be viewed as a compromise between the simplicity of pure static optimization and the rigour of optimal control methods such as direct collocation, which can provide more robust solutions at greater computational expense. CMC was favoured here for its speed and simplicity, and the existence of an established implementation in OpenSim.

Paired trajectories of states and excitations may be used to initialize policies, for example by imitation learning and DAgger \cite{ross2011reduction}, but this approach failed here, possibly due to a reliance of CMC upon  additional ‘residual’ ideal torque actuators to guarantee the success of optimization. However, the trajectory of states derived from CMC, which includes muscle fibre lengths and activation levels, was found to contain useful information for policy learning. Following \cite{peng2018deepmimic}, we used two methods to convey this information to the agent.\bigskip

\textbf{1. Reward shaping}. An imitation term was incorporated into the reward, describing the closeness of the agent’s kinematics (joint angles and speeds) to those of the reference kinematic trajectory at a given time step. The full reward function is described by:
\begin{equation}
	r_{t} = w_{1}*r_{t}^{imitation} + w_{2}*r_{t}^{goal}   
\end{equation} where $r_{t}^{imitation}$ is the imitation term and $r_{t}^{goal}$ describes the agent’s concordance with the target velocity vector at time \textit{t}. At each step, the motion clip used to compute the imitation objective was selected from the three available clips on the basis of minimum euclidean distance between the clip’s speed and the magnitude of the desired velocity vector. The choice of the coefficients $w_{1}$ and $w_{2}$ by which these separate terms were weighted in the overall reward scheme was found to be an important determinant of learning progress. Values of 0.7 and 0.3 for imitation and goal terms respectively were used in the final model. \bigskip

\textbf{2. Reference state initialization (RSI)}. At the start of each episode, a motion clip was selected at random and a single state was sampled and used for initialization of the agent. The sampled state index determined the reference kinematics used to compute imitation rewards, which were incremented in subsequent time steps. 

\subsubsection{Phase 2: Policy fine-tuning}
Following a period of training, the imitation objective was removed from the reward function, leaving a sole goal term for further training.

\subsection{Experiments and results}
Test scores in excess of 9700 were achieved after training for 10 hours - approximately 2.5m samples - on a 4-core machine (Intel core i7 6700). Further training improved the final score to 9853.

\begin{figure}
    \begin{center}
        
	\label{img:lance-ablations}
	\includegraphics[width=0.85\textwidth]{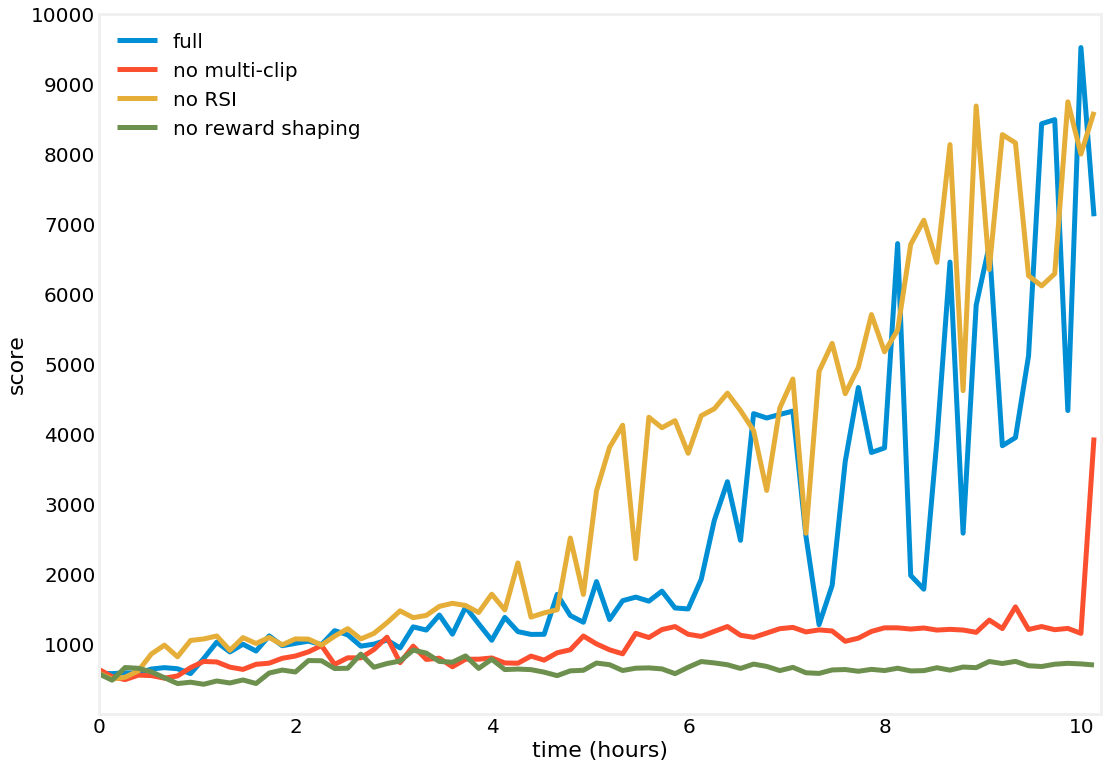}
	\caption{Performance impact of key feature ablations. Each line represents the average score over three independent runs. \textit{no multi-clip} refers to models trained using only a single motion clip at 1.25m/s. Agents trained without RSI scored relatively well, but did so by learning a policy that favoured remaining still.}
	\end{center}
\end{figure}

\subsection{Discussion}

Learned policies demonstrated natural walking behaviours and were capable of adapting both speed and direction on demand. Despite restriction of motion data to just three discrete speeds during training, agents learned to generalize to a continuous range of walking speeds, within and beyond the range of these clips, and were able to effectively combine changes in speed with changes of direction, for which no motion data were provided. Both reward shaping and reference state initialization proved critical for effective learning, with the absence of either leading to total decimation of performance. In a boost to the potential flexibility of the method, and unlike in \cite{peng2018deepmimic}, training was not dependent on the use of a synchronizing phase variable. 

To some extent, imitation resulted in suboptimal learning - for example, the goal term was based on a constant pelvic velocity, but there is considerable fluctuation in pelvic velocity during normal human walking. This may explain why a period of finetuning without imitation boosted scores slightly. Further improvements may have been possible with the use of data from turns during gait, which unfortunately were not available during the competition. Nevertheless, the techniques described here may find use in the rapid initialization of policies to serve as models of motor control or as the basis for the learning of more complex skills. Code, detailed design choices and hyperparameters may be viewed at \url{https://github.com/lancerane/NIPS-2018-AI-for-Prosthetics}.

\section{Accelerated DDPG with synthetic goals}\label{s:aditya}
\sectionauthor{Aditya Bhatt}

A Deep Deterministic Policy Gradient (DDPG)\citep{lillicrap2015continuous} agent is trained using an algorithmic trick to improve learning speed, and with the Clipped Double-Q modification from TD3 \cite{fujimoto2018addressing}. Transitions sampled from the experience buffer are modified with randomly generated goal velocities to improve generalization. Due to the extreme slowness of the simulator, data is gathered by many worker agents in parallel simulator processes, while training happens on a single core. With very few task-specific adjustments, the trained agent ultimately gets 8th place in the NeurIPS 2018 AI for Prosthetics challenge.

\subsection{Methods}

\subsubsection{Faster experience gathering}
Because running the simulator is very slow, depending on the different machines available, between 16 and 28 CPU cores were used to run parallel instances of the simulator, for the purpose of gathering data. The transitions were encoded as \((s,a,r,s')\) tuples with values corresponding to the state, the action, the reward and the next state. Transitions were sent into a single training thread's experience dataset. The simulator was configured to use a lower precision, which helped speed up execution. This risks producing biased data which could hurt agent performance, but no significant adverse impact was observed.

\subsubsection{Pre-processing}
The only augmentation done to the sensed observations was to change all absolute body and joint positions to be relative to the pelvis's 3-dimensional coordinates. As a form of prior knowledge, the relative coordinates ensure that the agent does not spend training time trying to learn that the reward is invariant to any absolute position.  The components of the 353-dimensional\footnote{Each observation provided by the simulator was a python dict, so it had to be flattened into an array of floats for the agent's consumption. This flattening was done using a function from the helper library \cite{seungjaeryanlee}. Due to an accident in using this code,  some of the coordinates were replicated several times, thus the actual vector size used in the training is 417.} state vector had very diverse numerical scales and ranges; however, no problem-specific adjustment was done to these.

\subsubsection{Algorithm}
Because the simulator was slow to run, on-policy algorithms like PPO were impractical on a limited computational budget. This necessitated using an off-policy algorithm like DDPG. The same neural network architecture as in the original DDPG paper was used, with the two hidden layers widened to $1024$ units. Batch normalization was applied to all layers, including the inputs; this ensured that no manual tuning of observation scales was needed.

DDPG and its variants can be very sample-efficient, however its sample-complexity is raised by artificially slowing down the learning with target networks (which are considered necessary to avoid divergence). To alleviate this problem, a new stabilizing technique\footnote{Unpublished ongoing research by the author, to be made public soon.} was used to accelerate the convergence of off-policy TD learning. This resulted in much faster learning than is usual.

A problem with doing policy improvement using Q functions is that of Q-value overestimation, which can cause frequent collapses in learning curves and sub-optimal policies. The Clipped Double Q technique was used to avoid this problem; this produced an \textit{underestimation} bias in the twin critics, but gave almost monotonically improving agent performance.

\subsubsection{Exploration}
Gaussian action noise caused very little displacement in position. It was also not possible to produce the desired amount of knee-bending in a standing skeleton with DDPG's prescribed Ornstein-Uhlenbeck noise, so instead another temporally correlated \textit{sticky gaussian} action noise scheme was tried: a noise vector was sampled from $\mathcal{N}(0, 0.2)$ and added to the actions for a duration of 15 timesteps.

\subsubsection{Reward shaping}
Aside from a small action penalty, the original reward function uses the deviation between the current and target velocities as: $$r=10 - ||v_{current} - v_{target}||^2$$This encourages the agent to remain standing in the same spot. Also, the reward does not penalize small deviations much. To provide a stronger learning signal, an alternative reward function was employed: $$r=\frac{10}{(1+||v_{current} - v_{target}||^2)}$$This change produces a stronger slope in the reward with a sharp peak at the desired velocity, ideally encouraging the agent to aim for the exact target velocity, and not settle for nearby values.

\subsubsection{Synthetic goals}
Despite having a shaped reward, the task is similar in nature to problems in RL with goals. For any transition triplet $(s,a,s')$, the reward $r$ can be directly inferred using the aforementioned function, because $s$ contains $v_{target}$ and $s'$ contains $v_{current}$ . Then, in a similar spirit to Hindsight Experience Replay \cite{Andrychowicz2017HindsightER}, whenever a batch of transitions is sampled, a batch of synthetic $v_{target}$ vectors with entries from $\mathcal{U}(-2.5, 2.5)$ is transplanted into $s$. The new $r$ is easily computed and the actual training therefore happens on these synthetic transitions.

The point of synthetic goals is that the agent can reuse knowledge of any previously attempted walking gait by easily predicting correct returns for completely different hypothetical goal velocities.

\subsection{Experiments}
In terms of hyperparameters, a batch size of 1024 was used. There was no frame-skipping. The optimizer was RMSprop with learning rates of $10^{-4}$ and $10^{-3}$ for the actor and critic respectively. The critic used a weight decay of strength $10^{-2}$. The batch size was 256. A discount factor of 0.98 was used.
A single training run contained 2 million simulator steps, with the same number of transitions stored in the experience replay memory.
The wall-clock time for a single training run was approximately 18 hours, by which point the total reward would have stabilized around 9850.

\begin{figure}[!h]
    \centering
    \includegraphics[width=0.8\textwidth]{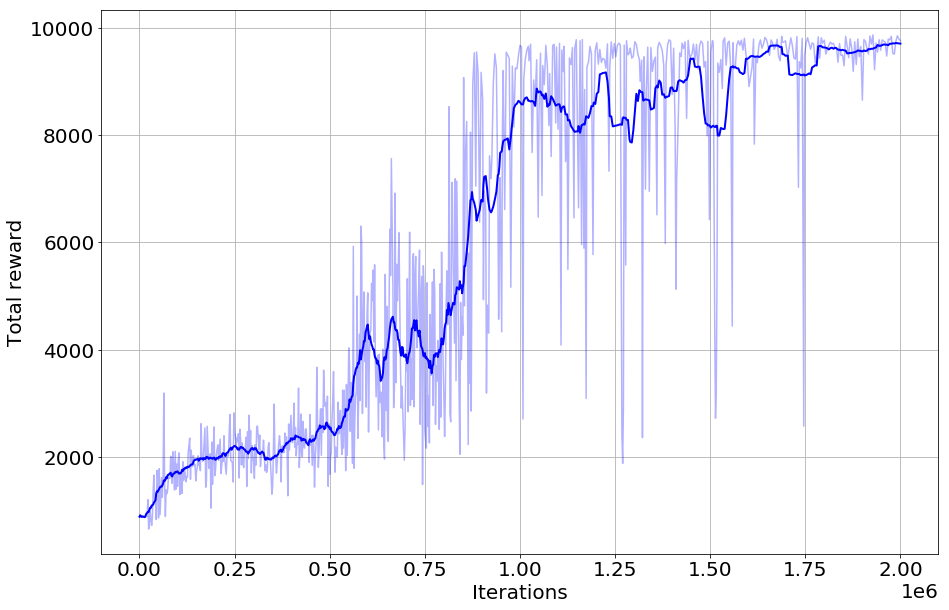}
    \caption{An example training run (with smoothing).}
    \label{fig:setup-penalty}
\end{figure}

It was noticed only a short while before the competition's deadline that the most difficult part of the episode and also the one where improvement would gain the most reward points was the first 150 steps, when the skeleton goes from standing to walking. To this end, a separate training run was launched to fine-tune the agent on the first phase of $v_{target}^x=1.25$, with an extra coordinate for the episode timestep added to the state vector. To save time, this agent started training with a union of the experience memories of three different good training runs, in addition to its own growing memory. In a short while, with a batch size of 1024, this agent performed on average better than all three of the agents. The performance on the start phase improved, with a minor degradation in the post-start phase.

Even better test-time performance was then extracted by using the critics, at each episode step, to choose the best action from 2000 candidates sampled from a Gaussian (of variance $0.2$) centered at the actor-produced action vector.
The final agent contained two network models: a start-phase fine-tuned network and a network from another good training run. This brought up the average score to 9912, with the exception of rare episode seeds with very difficult target velocities which caused agent to fall down.

During round 2, one of the hidden seeds corresponded to a particularly tricky velocity and the agent stumbled, bringing down the 10-seed average to 9852.

\subsection{Conclusion}
Even better performance could have been attained had the start-phase fine-tuned agent simply been trained for a longer time, and if problematic velocities had been emphasized in the training. That said, it is encouraging that a strong combination of algorithmic ingredients can competitively solve such a complex reinforcement learning problem, with very little problem-specific tailoring.

\section{Proximal Policy Optimization with improvements}\label{s:wangzhengfei}
\sectionauthor{Zhengfei Wang, Penghui Qi, Zeyang Yu, Peng Peng, Quan Yuan, Wenxin Li}

We apply Proximal Policy Optimization (PPO) \cite{schulman2017proximal} in the 
NeurIPS 2018: AI for Prosthetics Challenge. To improve the performance further, 
we propose several improvements, including reward shaping, feature engineering 
and clipper expectation. Our team placed 9th in the competition.

\subsection{Methods}

\subsubsection{Reward Shaping}
With substantial experiments with various combinations of observation and reward, 
we find that it seems hard to train the model successfully with a single reward. 
As in human walking in reality, we divide the whole walking procedure into 
phases and describe each phase with a reward function. We name these reward functions
as courses and our model is trained course by course. Details about the courses 
are shown in Table ~\ref{tb:wangzhengfei-courses}.

\begin{table}[h]
\centering
\caption{Courses for walking in AI for Prosthetics Challenge.}
\label{tb:wangzhengfei-courses}
 \begin{tabular}{c | c | c} 
 \hline
 Courses    & Reward Function Changes                       & Intuition \\
 \hline
 Penalty    & lean back and low pelvis height               & typical falling pattern, avoid early termination \\
 Init       & pelvis velocity and survival reward           & motivate agent to extend his leg to move forward \\
 Cycle      & velocity of two feet                          & move both legs in turns \\
 Stable     & distance between feet (minus, as punishment)  & avoid too much distance between two feet \\
 Finetune   & official evaluation (replace pelvis velocity) & adapt requested velocity for competition \\
 \hline
\end{tabular}
\end{table}

Near the end of the competition, we propose a new reward function based on 
exponential function, as shown below.  

\begin{equation}
  r_t = e^{|v_x(t) - tv_x(t)|} + e^{|v_z(t) - tv_z(t)|}
  \label{eq:wangzhengfei-reward}
\end{equation}
where $v(t)$ and $tv(t)$ represent current velocity and target (requested) 
velocity at step $t$. This function is smoother and provide larger gradient 
when there is small distance between current velocity and requested velocity.

\subsubsection{Clipped Expectation}

A requested velocity introduces stochasticity to the environment, and the agent 
always tries to adapt it to get higher reward. However, we find that 
when difference between current velocity and requested velocity is big 
enough, the agent will become unstable and perform worse. Also the episode 
will terminate early, resulting in score loss. To handle this problem, 
we manually set a threshold $Th$ for the agent, whose current velocity is 
$v$. We clip the requested velocity into the range $[v - Th, v + Th]$ before 
passing it to the agent.

\subsection{Experiments and results}

\subsubsection{Baseline Implementation}
We use PPO as our baseline. To describe the state of the agent, we apply 
feature engineering based on last year's solutions. To stress the importance 
of the requested velocity, we copy it twice in the state. 

To improve the performance of parallel computing, we replace default sampling 
environments from subprocess based with Ray \cite{moritz2018ray}. This 
makes our PPO scalable across servers and clusters. Furthermore, inspired by 
Cannikin Law, we propose to launch extra sampling environments to speedup. 
We have open-sourced the code
\footnote{\url{https://github.com/wangzhengfei0730/NIPS2018-AIforProsthetics}} 
and details about the states and parallel computing can be found online.

\subsubsection{Clipped Expectation}
During the competition's round 2 evaluation, there is one episode where our model 
fails to walk for complete episode all the time. Our best model without 
clipped expectation can achieve about 800 steps and get less 8000 points. 
We set a threshold for axis x as 0.3 and axis z as 0.15. This modification 
helps our model complete that episode and improve with nearly 1500 points. 

\subsubsection{Slow Start}
\begin{figure}[h]
    \begin{center}
        \includegraphics[width=0.7\textwidth]{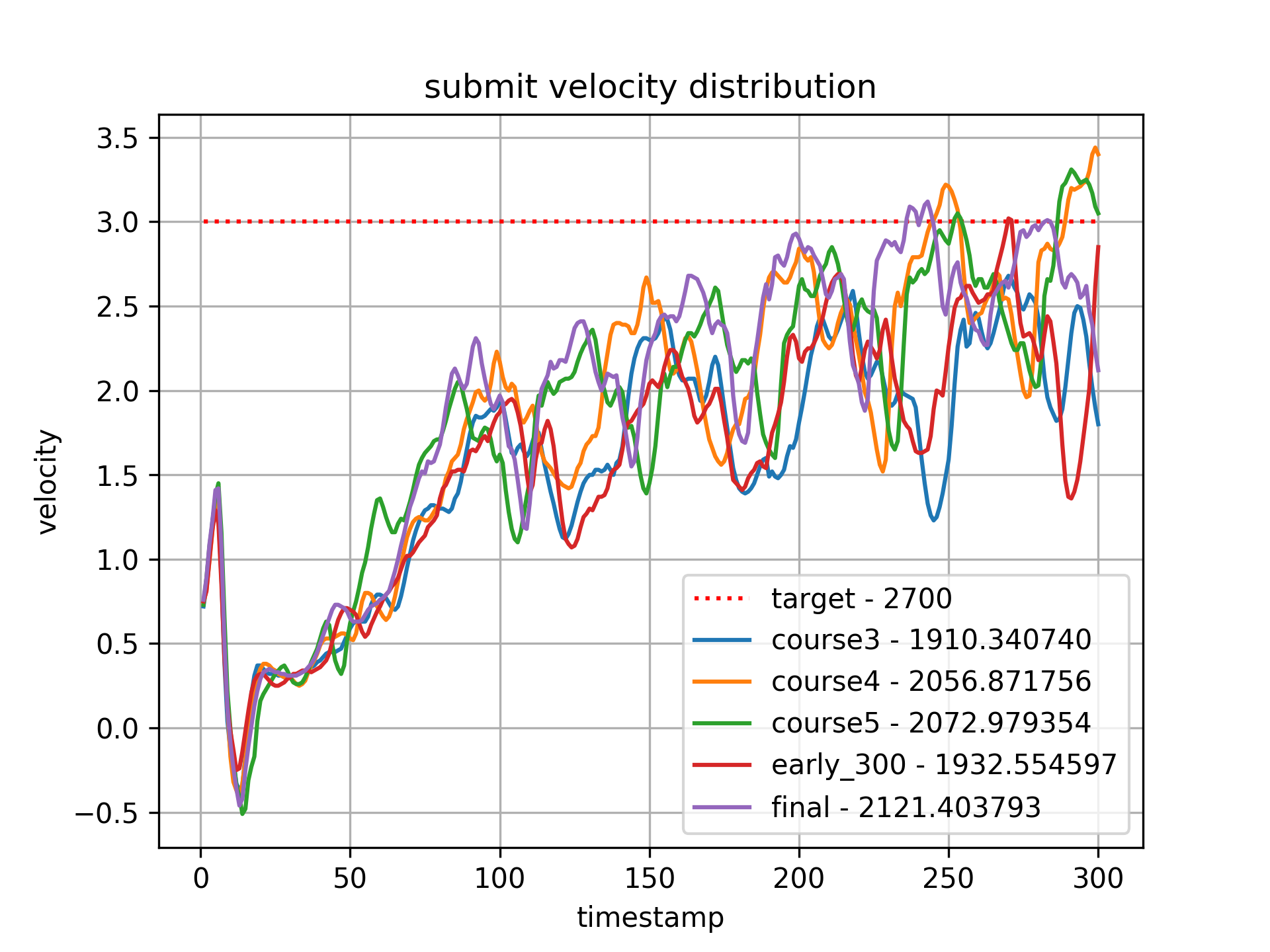}
    \end{center}
    \caption{Velocity distribution in round 1.}
    \label{fig:wangzhengfei-vd}
\end{figure}

We have plotted velocity distribution along the episode in round 1 for 
analysis, and we can regard round 1's requested velocity as 
$[3.0, 0.0, 0.0]$. The plot is shown in Fig.~\ref{fig:wangzhengfei-vd}, our model's 
velocity at first 50 steps is extremely slow, and even negative at times. We have tried 
several methods to fix this problem during both round 1 and 2. We think 
it could kind of overfitting and gave up on trying to fix it. 

\subsection{Discussion}

We apply PPO to solve the AI for Prosthetics Challenge, and to improve its 
performance further, we implement several modifications. As in 
real human walking, we divide the whole task into courses and train the 
agent course by course. We also propose several minor improvements 
to speed up and adapt better to the requested velocity. 

Our model has a very slow velocity at the beginning steps, which result 
in a nearly 100 points loss. However, we cannot afford retraining a new 
model and gave up on fixing this issue. Besides, our clipped expectation 
shall influence the performance for unnecessary clip. These could have 
been improved to some degree.

\section{Ensemble of PPO Agents with Residual Blocks and Soft Target Update}\label{s:rukia}
\sectionauthor{Yunsheng Tian, Ruihan Yang, Pingchuan Ma}

Our solution was based on distributed Proximal Policy Optimization \cite{schulman2017proximal} for stable convergence guarantee and parameter robustness. In addition to careful observation engineering and reward shaping, we implemented residual blocks in both policy and value networks and witnessed faster convergence. To address the instability of gait when the target speed changes abruptly in round 2, we introduced Soft Target Update for a smoother transition in observation. We also found Layer Normalization helps to learn, and SELU outperforms other activation functions. Our best result was established on multiple agents fine-tuned at different target speeds, and we dynamically switch agent during evaluation. We scored 9809 and placed 10th in the NeurIPS 2018 AI for Prosthetics competition.

\subsection{Methods}

\subsubsection{Observation Engineering}

In our task, the full state representation of the dynamic system is determined by hundreds of physical quantities which are very complex and not efficient for the agent to learn. Therefore, we proposed several observation engineering techniques to alleviate this issue.

\textbf{Dimension Reduction}. Among the original observation provided, we carefully observed each kind of physical quantities of the skeleton model and found that acceleration-related values have a much larger variation range than others and seem to be unstable during simulation. Considering position and velocity quantities are reasonably enough to represent model dynamics, we removed acceleration values from the observation. As a result, we found the removal did not deteriorate performance and even sped up convergence due to the reduction of nearly 100 dimensions on network input.

\textbf{Observation Generalization}. Because our ultimate goal is to walk at a consistent speed, there is no need for the agent to care about its absolute location, i.e., the agent's observation should be as similar as possible when walking in the same pose at different places. Therefore, we subtract the position of all bodies, except the pelvis, by the pelvis position in observation. Doing so prevents many input values to the policy and value networks from going to infinity when the agent runs farther and farther, but this still lets the agent know the current distance from the starting point.

\textbf{Soft Target Update}. From our observation, our trained agents are more likely to fall during abrupt changes of target velocity in observation. The lack of generalization of our policy network could be the reason of this falling behavior, but it is also possible that our agent reached a local optima by falling down in the direction of the changed target velocity. Thus, denoting the target velocity that we feed into observation as $v_{curr}$, and the real target velocity as $v_{real}$, we smoothed the change of $v_{curr}$ by linear interpolation between $v_{curr}$ and $v_{real}$ in each step: $v_{curr}=\tau*v_{curr}+(1-\tau)*v_{real}$, where $\tau$ is a coefficient between 0 and 1 controlling the changing rate of target velocity. In practice we choose $\tau=0.8$ which guarantees $v_{curr}\approx v_{real}$ within 20 steps.

\subsubsection{Reward Shaping}

Our reward function consists of three parts, shown in equation \ref{eq:rukia_reward}.
\begin{equation}
\begin{split}
    Reward & =r_{speed}+r_{straight}+r_{bend}\\
    & =w_{speed}*||\max(\sqrt{|\Vec{v}_{target}|}-\sqrt{|\Vec{v}_{pelvis}-\Vec{v}_{target}|},\Vec{0})||_1\\
    & +w_{straight}*\sum_{i=head,torso}{(\frac{||\Vec{v}_{target}\times\Vec{v}_{i}||_2}{||\Vec{v}_{target}||_2})^2}\\
    & +w_{bend}*\sum_{i=left\_knee,right\_knee}\min(\max(\theta_{i},-0.4),0),
\end{split}
\label{eq:rukia_reward}
\end{equation}
where $w_{speed},w_{straight},w_{bend}$ are weights chosen as $5,4,2$ respectively, $\Vec{v}$ represents 2-dimension velocity vector on the X-Z plane, and $\theta$ is a negative value that accounts for the bending angle of a knee. The detailed meaning of each part of the reward is discussed below.

\begin{figure}
    \centering
    \begin{subfigure}{0.45\textwidth}
        \includegraphics[scale=0.45]{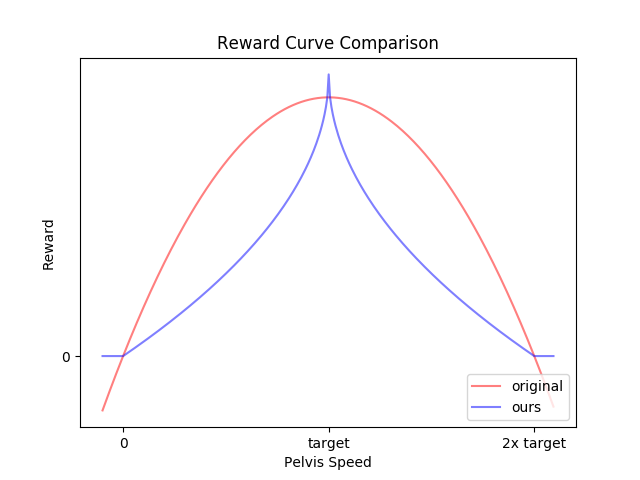}
        \caption{The reward curve comparison between the original reward function and our version.}
        \label{fig:rukia_rew_speed}
    \end{subfigure}
    \begin{subfigure}{0.45\textwidth}
        \centering
        \includegraphics[scale=0.45]{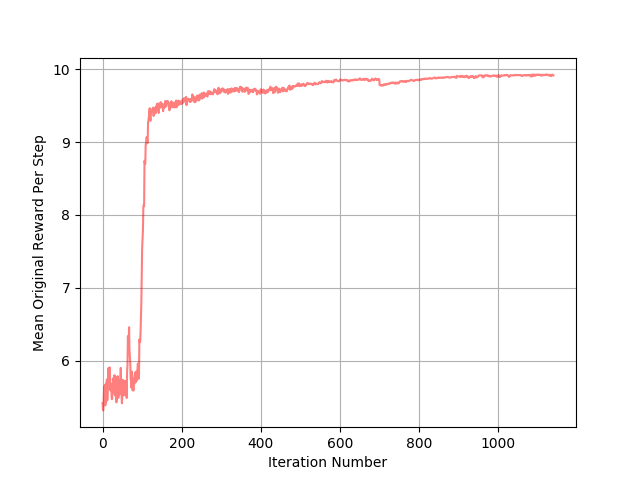}
        \caption{An example learning curve from final agents.}
        \label{fig:rukia_result}
    \end{subfigure}
    \caption{Modified reward function comparison and an example overall learning curve}
\end{figure}

\textbf{Speed matching}. Obviously, it is most important that the velocity of the pelvis matches the target velocity. In practice, we found it is easy for the agent to speed up but hard to control its speed around the target value. Thus, instead of the speed term in the original reward which is $-||\Vec{v}_{pelvis}-\Vec{v}_{target}||_2$, we changed the L2-norm to squared root for more sensitivity in the region near the target, which turned out to converge faster than the original reward (see figure \ref{fig:rukia_rew_speed}). But, only using speed matching reward seemed insufficient for this task due to some local optima in learning gaits, so we also introduced other auxiliary reward terms that helped the agent behave more reasonably.

\textbf{Going straight}. Because the speed matching reward only pays attention to the agent's pelvis, sometimes the agent cannot keep walking straight even though the movement of its pelvis point nearly matches the target value. Thus, we also encourage its head and torso to move at the target speed, which further ensures that the skeleton body keeps vertically straight.


\textbf{Bending knees}. Our agents could hardly learn to bend its knees before adding this reward term. Also, keeping the legs straight makes the agent more likely to fall. This term encourages the agent to bend its knee to a small angle which improves its stability of walking at a consistent speed. 

\subsubsection{Residual Blocks}

\begin{wrapfigure}{r}{5cm}
    \centering
    \includegraphics[width=6cm]{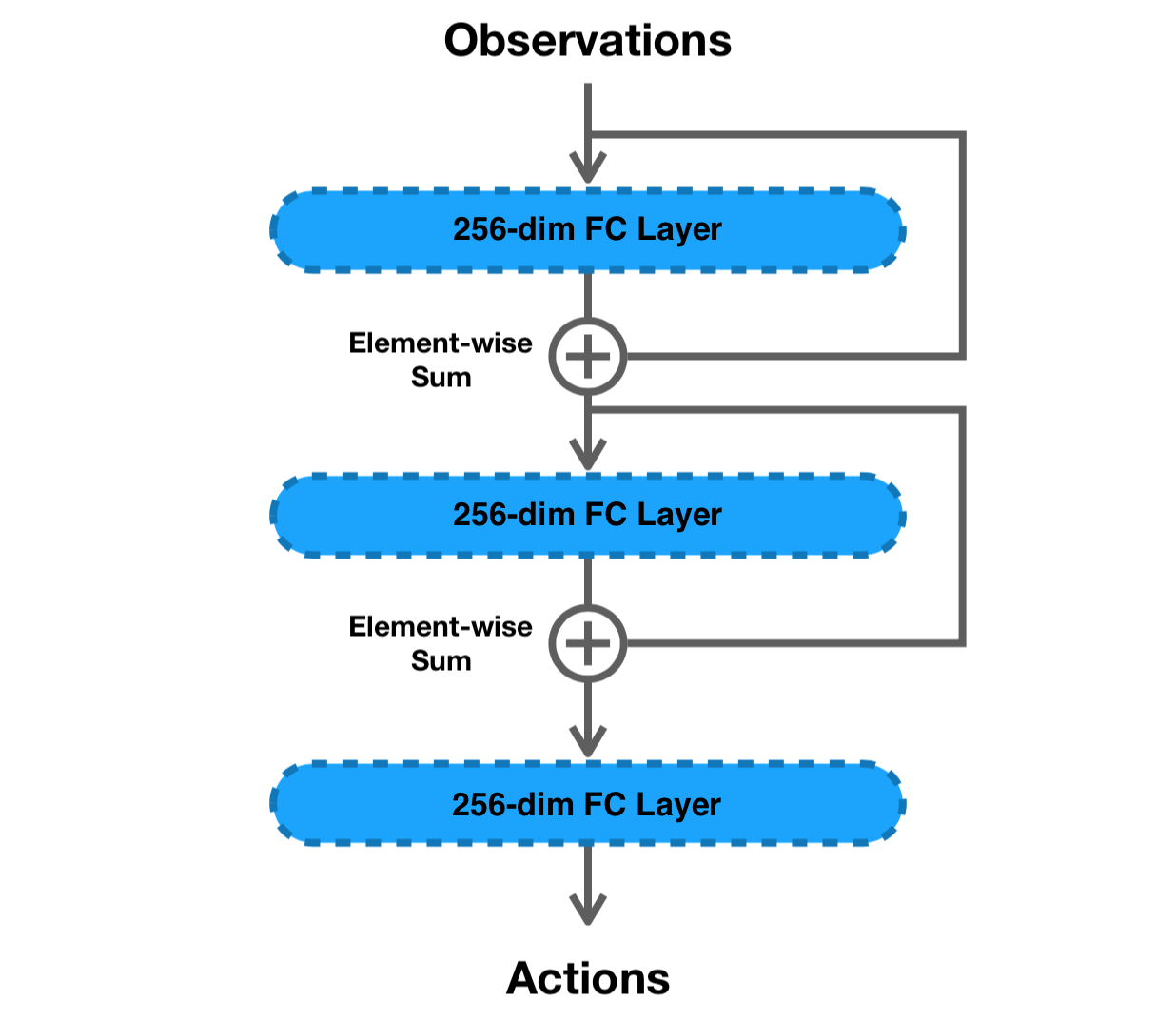}
    \caption{Policy Network Overview}
    \label{fig:rukia_network}
\end{wrapfigure}

We applied the idea of residual blocks to our policy network and value network, i.e., we added shortcut connections on the basis of 4 fully connected layers, as illustrated in figure \ref{fig:rukia_network}. Consequently, it further improved about 10\% on our shaped reward and sped up convergence compared to results based on networks without shortcut connections.

\subsubsection{Ensemble of Agents}

Limited by our time and computing resources, we found that it was hard to train an agent that automatically adapts to a different speed, so we trained several agents that specialize at one particular target velocity and combined them at evaluation time. To select target velocities for training, we generated thousands of different targets using the original generation algorithm provided, and observed that target velocities approximately conformed to Gaussian distributions $v_{x}\sim\mathcal{N}(1.27,0.32)$ and $v_{z}\sim\mathcal{N}(0, 0.36)$. We then picked $v_{x}$ from \{0.7, 0.9, 1.25, 1.4, 1.6, 1.8\} which are within the $2\sigma$ range of the distribution and simply used $v_{z}=0$ to train agents. Consequently, it was a clear boost on performance but not an elegant and scalable approach compared to a real multi-speed agent.

\subsubsection{Additional Tricks}

Based on our benchmark experiments, we found adding Layer Normalization before activation stabilizes the convergence and SELU is better than other activation functions, such as RELU and ELU. We also tried several scales of parameter noise, but it turned out to only have a minor effect on results, thus we did not add noise.

\subsection{Experiments and results}

We ran experiments mainly on the Google Cloud Engine and the Tianhe-1 Supercomputer (CPU: Intel Xeon X5670). Our final result took us one day of training on 240 CPUs on Tianhe-1. Our PPO implementation was based on OpenAI's Baselines\citep{baselines} and achieved parallelization by distributed sampling on each process.

The hyper-parameters for our reward function and some tricks are stated above. For hyper-parameters of RL algorithm and optimizer, we mainly used default values from OpenAI's PPO implementation and the Adam optimizer with some minor changes. We sampled 16000 steps per iteration, used a batch size of 64, a step size of $3*10^{-4}$, an epoch of 10 for optimization per iteration, annealed clipping $\epsilon$ of 0.2, and policy entropy penalization of 0.001. For more detailed information, please see our source code at \url{https://github.com/Knoxantropicen/AI-for-Prosthetics}. 

Figure \ref{fig:rukia_result} shows an example learning curve of an agent adapted to running at 1.6 m/s. The Y-axis of the figure represents mean original reward per step (which is not our shaped reward, and 10 is the maximum). It took more than 16M samples to get this result. The cumulative reward of a whole trajectory usually varies between 9800 and 9900 according to different random seeds. We achieved 9809 in the official evaluation.

\subsection{Discussion}

Thanks to all of the team members' effort, we got a pretty good result and a satisfying rank. However, our solution is relatively brute-force and naive compared with those winning solutions and could be improved in multiple ways. So, this section focuses on potential improvements.

\subsubsection{Better Sample Efficiency using Better Off-policy Methods}
Our solution is based on Proximal Policy Optimization (PPO), whose sample efficiency is much better than previous on-policy methods. Nevertheless, recent advances in off-policy methods like DDPG, SAC, TD3 has shown us that off-policy methods sometimes could perform better in continuous control tasks. Due to the time-consuming simulation in this challenge, methods with better sample efficiency could be a better choice.

\subsubsection{Special Treatment to Crucial Points}
    According to our reward analysis, most of our reward loss comes from two parts: 
    \begin{enumerate}
        \item [-] \textbf{Starting stage}.
            Our agent suffers from a slow start. Perhaps a specialized starting model could improve the agent's performance in the starting stage.
        \item [-] \textbf{Significant direction change}. Currently, we ensemble multiple models and use the soft update of target velocity to deal with the change of target velocity, but using the ensemble of models trained with different target velocity is likely to achieve a sub-optimal solution. Instead, the ensemble of models for different specialized tasks, like changing direction and running forward/backward, could be a better solution.
    \end{enumerate}
 Moreover, our agent performs extremely poorly in some rare situation. For instance, if the target velocity is extremely slow, our agent is still likely to go forward at a high speed and is unable to remain still. Maybe some special treatment to these corner cases could also help our agent.

\subsubsection{Model-based Solution}
In this challenge, complex physical simulation in high dimensional continuous space makes sampling pretty time-consuming. Another alternative is to use model-based methods to get more imaginary samples for training, and a known reward function in this challenge makes model-based methods feasible. 

\subsubsection{Observation with False Ending Signal}
In this kind of infinite-horizon tasks, it is natural to set a maximum time step limit for a simulation trajectory. Thus, the sample from the last step is associated with an ending signal, but the several previous samples are not, even if their states are quite similar. When the RL algorithm considers this ending signal for computing and predicting Q/V values, e.g., in 1-step TD estimation, this difference on ending signal could cause a significant error in value prediction, which destabilizes the training process. In our implementation, this situation is not well treated, though a one-step bootstrap could solve it.
\section{Collaborative Evolutionary Reinforcement Learning}\label{s:shawk91}
\sectionauthor{Shauharda Khadka, Somdeb Majumdar, Zach Dwiel, Yinyin Liu, Evren Tumer}

We trained our controllers using Collaborative Evolutionary Reinforcement Learning (CERL), a research thread actively being developed at Intel AI. A primary reason for the development and utilization of CERL was to scale experimentation for Deep Reinforcement Learning (DRL) settings where interacting with the environment is very slow. This was the case with the Opensim engine used for the AI for Prosthetics challenge. CERL is designed to be massively parallel and can leverage large CPU clusters for distributed learning. We used Intel Xeon servers to deploy the CERL algorithm for learning in the osim environment.

\begin{figure}[h]
\centering
\includegraphics[width=0.75\textwidth]{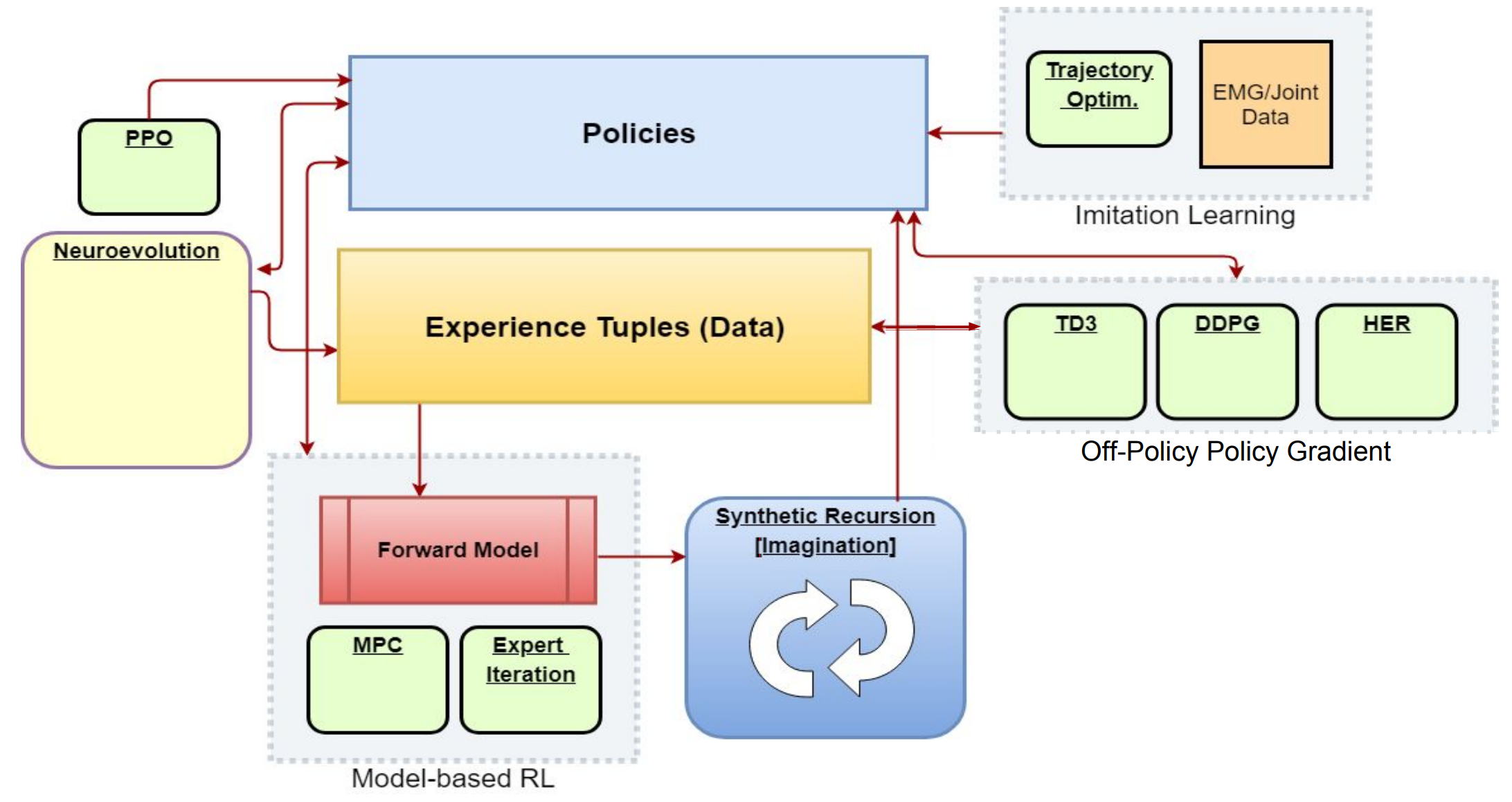}
\caption{Illustration of the Collaborative Evolutionary Reinforcement Learning framework}
\label{cerl}
\vspace{-2em}
\end{figure}

\subsection{Methods}\label{sss:distributed-shawk}

Collaborative Evolutionary Reinforcement Learning (CERL) is a multilevel optimization framework that leverages a collection of learners to solve a Reinforcement Learning (RL) problem. Each learner explores independently, while exploiting their collective experiences jointly through the sharing of data and policies during learning. A central integrator (neuroevolution) serves to drive this process by adaptively distributing resources - imposing a strong selection pressure towards good learners. The ‘collective learner’ that emerges from this tightly integrated collaborative learning process combines the best of its composite approaches. There are five core components that jointly describe the CERL framework:

\begin{enumerate}
    \item Data Generators - Generate experience by rolling out in the environment.
\item Data Repository - Store collective experiences.
\item Data Consumers - Exploit collective data to learn policies.
\item Policy Repository - Store policies.
\item Integrator - Exploit decomposability of policy parameters to synergestically combine a diverse set of policies, and adaptively redistribute resources among learners.
\end{enumerate}

 A general flow of learning proceeds as follows: a group of data generators run parallel rollouts to generate experiences. These experiences are periodically pushed to the data repository. A group of data consumers continuously pull data from this database to train policies. The best policy learned by each data consumer is then periodically written to the policy repository. The data generators and consumers periodically read from this policy database to update their populations/policies, hereby closing the learning loop. The integrator runs concurrently exploiting any decomposability within the policy parameters space and operates in combining the diverse set of behaviors learned. The integrator also acts as a 'meta-learner' that adaptively distributes resources across the learners, effectively adjusting the search frontier.

Figure \ref{cerl} depicts the organization of the CERL framework. A crucial feature of the CERL framework is that a wide diversity of learning algorithms spanning across the off-policy, on-policy, model-free and model-based axis of variation can work collaboratively, both as data generators and consumers. Collaboration between algorithms is achieved by the exchange of data and policies mediated by the data and policy repository, respectively. The \textbf{"collective learner"} that emerges from this collaboration inherits the best features of its composite algorithms, exploring jointly and exploiting diversely.
\vspace{-1em}

\subsection{Experiments and Results}

\textbf{Reward Shaping: } A big advantage of using neuroevolution as our integrator was that we could perform dynamic reward shaping on the entire trajectory originating from a controller. This allowed us to shape behaviors at the level of gaits rather than static states. Some target behaviors we shaped for were bending the knee (static) and maximizing the swing for the thighs (dynamic). This really helped train our controllers in running and matching the 3m/s speed in the first round with a semi-realistic looking gait. 

We started the second round (adaptive walking) by seeding the best policy from the first round. However, this was perhaps our biggest error. Unlike the first round, the second round required our agent to move at a much slower speed. The gait with bent knees and a hurling movement pattern that our agent learned in the first round was not ideal for the second round. Our bootstrapped agent learned quickly at first but hit a ceiling at ~9783 (local evaluation). The agile and rather energetic movement pattern that our agent implemented from Round 1 was counterproductive for Round 2. This pattern led to high overshoot and twitching (jerk) of the waist. This was exacerbated for slower target speeds and led to large losses especially for our 20Hz controller (frameskip=5).

\textbf{Hardware:} The biggest bottleneck for learning was experience generation, stemming from the high-fidelity of the OpenSim engine. To address this issue we leveraged the parallelizability of CERL and used a Xeon server (112 cpu cores) in addition to a GPU to train our controller.

\vspace{-1em}
\subsection{\textbf{Discussion}}
We used CERL to develop controllers for the AI for Prosthetics challenge. The core idea behind the implementation of CERL was to leverage large CPU nodes in scaling DRL for settings where interacting with the environment is slow and laborious. We leveraged a diversity of reinforcement learning algorithms to define a collaborative learner closely mediated by the the core integrator (neuroevolution). The challenge provided a very good platform to test some exploratory ideas, and served to accelerate the development the CERL framework which is an active research effort at Intel AI. Future work will continue to develop and expand the CERL framework as a general tool for solving deep reinforcement learning problems where interactions with the environment is extremely slow.

\section{Asynchronous PPO}\label{s:HP}
\sectionauthor{Jeremy Watson}

\graphicspath{{../img2018/90_jeremy/}} 

We used a vanilla implementation of PPO in the open-source framework Ray RLLib \citep{ppo,moritz2018ray}.  Highest position achieved was a transitory top-10 in an intermediate round (subsequently falling to 17) but no final submission was made due to issues with action space.  Some minor reward shaping was used, along with quite a lot of hardware - typically 192 CPUs, but no GPUs, over several days per experiment.  The model achieved comparatively poor sample efficiency - best run took ~120 million steps in training, somewhat more than the Roboschool Humanoid environments  which were trained on 50-100 million timesteps in\citep{ppo}.  Initially we used the baselines \citep{baselines} PPO implementation and verified training with Roboschool which is much faster than Opensim \citep{seth2018opensim} but lacks physiologically realistic features like muscles. Source for Ray version on github\footnote{\url{https://github.com/hagrid67/prosthetics_public}}.

\subsection{Methods}

\subsubsection{Vanilla PPO with continuous actions}
Action $a_i$ is the sum of the $\tanh{}$ activations from the final hidden layer outputs $x_j$ with the addition of an independent Normal random variable with trainable logstdev $\sigma_i$:

$$ a_i := \sum_j W_{ij} x_j + b_i + Y_i, \ \ Y_i \sim N(0,\exp(\sigma_i))
$$

For submission and testing we set $\sigma_i=-10$ (typical trained values were around $-1$) to reduce exploration noise.  (In the baselines implementation a \texttt{stochastic} flag is available to remove the noise but this is not yet implemented in Ray.)

\subsubsection{Unfair advantage from actions outside the permitted space}

Initially an unbounded action space was used, as per vanilla PPO above.  It appears the model exploited this, using actions (muscle activations) greater than one.  This gave unfair advantage in the environment, achieving a score of 9,790 out of 10,000, which was briefly in the top 10, without any significant innovation.  Subsequently with bounded actions the model trained poorly.  This was not resolved and hence no final submission was made.
We conjecture that the slow timesteps are perhaps associated with the out-of-bounds actions, ie muscle activations $> 1$.

\subsubsection{Asynchronous Sampling}
This options "sample\_async" and "truncate\_episodes" within Ray allows episodes to continue running from one SGD epoch to the next, so that the SGD doesn't need to wait for longer-running episodes.  This allows all the environments to keep busy generating samples.

\subsubsection{Change of Opensim Integrator to EulerExplicit}
With the default configuration of the Runge-Kutta-Merson occasionally a single step would take tens of minutes, while most would take less than 1 second.  The Opensim C++ package was recompiled with the alternative integrator to give a performance boost as described elsewhere.  Along with a reduced accuracy setting, from $5\times10^{-5}$ to $10^{-3}$, this also reduced the huge variance in the step time.  

The ray asynchronous setting did not seem to be able to cope with long individual timesteps.  Attempts were made to accommodate the long timesteps but this was abandoned.

\begin{figure*}[ht!]%
        \centering
        \begin{subfigure}[t]{0.45\textwidth}
                \includegraphics[width=1\textwidth]{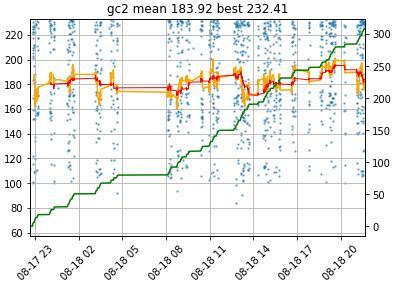}
                \label{fig:episode_gaps}
        \end{subfigure}
        \caption{Sporadic pauses in episode completions due to occasional slow timesteps under Runge Kutta Merson integrator (blue dots are episodes, the x-axis is the wall clock formatted as Month-Day-Hour. Green is completed timesteps.  Rolling means in yellow and orange.  Taken from round 1.  This implementation based on Baselines, not Ray. Here a single long-running episode holds up all the environments - the large gap represents about 3 hours. )}
        \label{fig:perf1}      
\end{figure*}



\begin{figure*}[ht!]%
        \centering
        \begin{subfigure}[t]{0.45\textwidth}
                \includegraphics[width=1\textwidth]{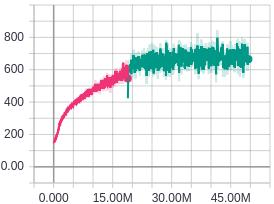}
                
        \end{subfigure}
        \begin{subfigure}[t]{0.45\textwidth}
                \includegraphics[width=1\textwidth]{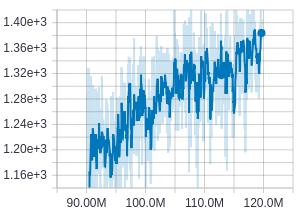}
                
        \end{subfigure}
        \caption{Performance with bounded vs unbounded action space (modified reward scales).  Although the unbounded chart is incomplete, it indicates that there was still improvement after 100m steps}
        \label{fig:perf1}      
\end{figure*}

\subsubsection{Binary Actions, Logistic function }
To confine the action space, "binary actions" $ a_i \in \{0,1\} $ were tried.  In Ray this was implemented as a "Tuple of Discrete spaces".  In our attempts it wasn't clear that this approach was learning any faster than continuous actions (as other competitors found). 

We also faced the difficulty that it would require a code change in Ray to use binary actions without exploration noise.  Un-normalised log-probabilities are used for the different discrete actions, and there is no explicit field for the log-stdev, which we were able to edit in the model snapshot file, in the continuous-action case.

We also tried applying the logistic function (with a factor of 10), $a_i=\frac{1}{1+e^{-10x}}$, to give similar behaviour, but using the established continuous action space output.  The factor was to steepen the gradient between 0 and 1 compared to the standard logistic.  By using a continuous action space  we could use the existing "DiagGaussian" distribution for the output space in training and then reduce the variance for submission.  (OpenAI baselines\citep{baselines} PPO has a \texttt{stochastic} flag but this is not implemented in Ray.

\begin{figure*}[ht!]%
        \centering
        \begin{subfigure}[t]{0.45\textwidth}
                \includegraphics[width=1\textwidth]{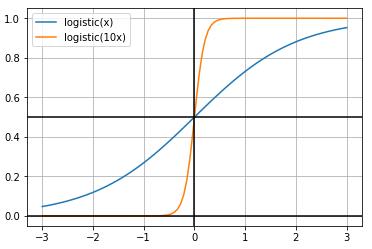}
                
        \end{subfigure}
        \begin{subfigure}[t]{0.45\textwidth}
                \includegraphics[width=1\textwidth]{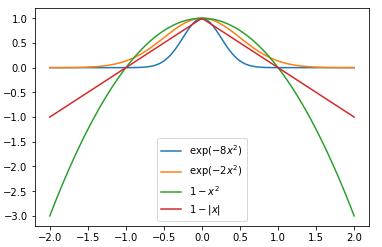}
                
        \end{subfigure}
        \caption{logistic function applied to actions, and some reward / penalty functions tried for reward shaping.)}
        \label{fig:logistic}      
\end{figure*}

\subsubsection{Observation Space}
We used all the 350 available observation scalars (not all these necessarily vary) plus the two target velocity values $\tilde{v}_{x,z}$.  We did not repeat the previous observation as velocities and accelerations were already available.  We made all $x,z$ positions relative to the pelvis, but left $y$ positions untouched.  We did not adjust velocities, accelerations, orientations, or joint angles.  Although muscle activations, fibre forces etc seem to correspond closely with the action outputs of the policy, we found slightly worse performance without them, so we kept them in.

\subsubsection{Basic reward shaping}
The default reward was $ b - ((v_x-\tilde{v}_x)^2 + (v_y-\tilde{v}_y)^2) $ where target velocity $\tilde{v}_{x,z}$ is initially $(1.25, 0)$ and subsequently varies.  With the default $b=10$ the agent tended to learn to simply stand still, to easily earn the reward for prolonging the episode, so we reduced $b$ to $2$ so that the velocity penalty could outweigh the base reward, leading to negative reward for standing still.

To discourage some sub-optimal gaits like walking sideways, we applied a penalty on pelvis orientation, $k_{pelvis}(r_x^2+r_y^2+r_z^2)$.  This was successful in causing the agent to walk facing forward. (The angular orientation (0,0,0) corresponded to facing forward.  We didn't get as far as adapting this penalty to the changing target velocity.)

\subsection{Unsuccessful experiments}

\subsubsection{leg-crossing penalty}

The agent developed a gait where its legs crossed in a physically impossible configuration. (The environment did not implement collision detection for the legs so they pass through each other.)  To discourage this we implemented a penalty on "hip adduction" - essentially the left-right angle\footnote{joint\_pos hip\_l [1] in the observation dictionary } of the thigh in relation to the hip.  A positive value means the thigh has crossed the centre line and is angled toward the other leg; a negative value means the thigh is pointing out to the side (we ignored this).  (The rotation of the thigh joint did not vary in the prosthetics model.)
The penalty was $k_{hip} (\theta_{hip} + 0.2)^+ $; this failed to cure the leg-crossing.  $k_{hip} \in \{0, 0.5, 1\} $

\subsubsection{Straight knee penalty}
The agent walked with straight legs.  Intuitively this would make it harder for the agent to respond to changes of target velocity.  We applied a penalty of  $k_{knees} (\theta_{knee} + 0.2)^+, k_{knees} \in \{0, 1, 2\} $  but were not able to detect any meaningful effect in the gait.

\subsubsection{Kinetics Reward / Imitation Learning}
An approach similar to DeepMimic \citep{peng2018deepmimic} was attempted, at a late stage after the bounds were imposed on the action space, with credit to Lance Rane (see \ref{s:lance} above), who suggested it on the public forum.  Sadly we were not able to train a walking agent using this approach.

\subsubsection{Wider network}
We found that increasing the hidden-layer width from 128 to 256 slowed down learning so we kept a width of 128 (despite an observation size of 352).  

\subsubsection{Frameskip}
We found no improvement in learning from repeating the actions for a few timesteps (2 or 3).

\subsection{Experiments and results}

\begin{table}
\caption{Hyper-parameters used in the experiments}
\begin{tabularx}{\columnwidth}{r|X}
  \toprule
  Actor network architecture & $[FC128, FC128]$, Tanh for hidden layers, \\
  Action noise & Normal / DiagGaussian with trainable variance (as logstdev).  PPO Baselines standard for continous actions. \\
  Value network architecture & $[FC128, FC128]$, Tanh for hidden layers and linear for output layer\\
  learning rate (policy and value networks) & 5e-5 (Ray PPO default), sometimes 1e-4 \\
  SGD epochs & 30 (Ray PPO default), sometimes 20 \\
  Batch size & 128 \\
  $\gamma$ & 0.99 \\
 \bottomrule
\end{tabularx}
\label{tab:hyperopt}
\end{table}

\section{Affiliations and acknowledgments}\label{s:acknowledgements}

\textbf{Organizers:} {\L}ukasz Kidzi\'nski, Carmichael Ong, Jennifer Hicks and Scott Delp are affiliated with Department of Bioengineering, Stanford University. Sharada Prasanna Mohanty, Sean Francis and Marcel Salathé are affiliated with Ecole Polytechnique Federale de Lausanne. Sergey Levine is affiliated with University of California, Berkeley.

\textbf{Team Firework, 1st Place, Section~\ref{s:BaiDu_NLP}}. Bo Zhou, Honghsheng Zeng, Fan Wang, Rongzhong Lian, is affiliated with Baidu, Shenzhen, China. Hao Tian is affiliated with Baidu US.

\textbf{Team NNAISENSE, 2nd Place, Section~\ref{s:nnaisense}}. Wojciech Jaśkowski, Garrett Andersen, Odd Rune Lykkebø, Nihat Engin Toklu, Pranav Shyam, and Rupesh Kumar Srivastava are affiliated with NNAISENSE, Lugano, Switzerland.

\textbf{Team JollyRoger, 3rd Place, Section~\ref{s:dqec}}. Sergey Kolesnikov is affiliated with DBrain, Moscow, Russia; Oleksii Hrinchuk is affiliated with Skolkovo Institute of Science and Technology, Moscow, Russia; Anton Pechenko is affiliated with GiantAI.

\textbf{Team Mattias, 4th Place, Section~\ref{s:mattias}}. Mattias Ljungström is affiliated with Spaces of Play UG, Berlin, Germany.

\textbf{Team ItsHighNoonBangBangBang, 5th Place, Section~\ref{s:apexddpg}}. Zhen Wang, Xu Hu, Zehong Hu, Minghui Qiu, Jun Huang are affiliated with Alibaba Group, HangZhou, China.

\textbf{Team jbr, Place 6th, Section~\ref{s:jbr}}. Aleksei Shpilman, Ivan Sosin, Oleg Svidchenko and Aleksandra Malysheva are affiliated with JetBrains Research and National Research University Higher School of Economics, St.Petersburg, Russia. Daniel Kudenko is affiliated with JetBrains Research and University of York, York, UK.

\textbf{Team lance, 7th Place, Section~\ref{s:lance}}. Lance Rane is affiliated with Imperial College London, London, UK. His work was supported by the NCSRR Visiting Scholars' Program at Stanford University (NIH grant P2C HD065690) and the Imperial College CX compute cluster.

\textbf{Team AdityaBee, 8th Place, Section~\ref{s:aditya}}. Aditya Bhatt is affiliated with University of Freiburg, Germany.

\textbf{Team wangzhengfei, Place 9th, Section~\ref{s:wangzhengfei}}. Zhengfei Wang is affiliated with inspir.ai 
and Peking University, Penghui Qi, Peng Peng and Quan Yuan is affiliated with inspir.ai, Zeyang Yu is affiliated with 
inspir.ai and Jilin University, Wenxin Li is affiliated with Peking University.

\textbf{Team Rukia, 10th Place, Section~\ref{s:rukia}}. Yunsheng Tian, Ruihan Yang and Pingchuan Ma is affiliated with Nankai University, Tianjin, China.

\textbf{Team shawk91, Place 16th, Section~\ref{s:shawk91}}. Shauharda Khadka, Somdeb Majumdar, Zach Dwiel, Yinyin Liu, and Evren Tumer are affiliated with Intel AI, San Diego, CA, USA.

The challenge was co-organized by the Mobilize Center, a National Institutes of Health Big Data to Knowledge (BD2K) Center of Excellence supported through Grant U54EB020405. The challenge was partially sponsored by Nvidia who provided four GPUs Titan V for top solutions, by Google Cloud Services who provided 70000 USD in cloud credits for participants, and by Toyota Research Institute who funded one travel grant.


\begin{thebibliography}{10}

  \providecommand{\url}[1]{{#1}}
\providecommand{\urlprefix}{URL }
\expandafter\ifx\csname urlstyle\endcsname\relax
  \providecommand{\doi}[1]{DOI~\discretionary{}{}{}#1}\else
  \providecommand{\doi}{DOI~\discretionary{}{}{}\begingroup
  \urlstyle{rm}\Url}\fi

\bibitem{Andrychowicz2017HindsightER}
Andrychowicz, M., Wolski, F., Ray, A., Schneider, J., Fong, R., Welinder, P.,
  McGrew, B., Tobin, J., Abbeel, P., Zaremba, W.: Hindsight experience replay.
\newblock In: NIPS (2017)

\bibitem{recurrent2018}
authors, A.: Recurrent experience replay in distributed reinforcement learning.
\newblock https://openreview.net/pdf?id=r1lyTjAqYX  (2018)

\bibitem{bahdanau2014neural}
Bahdanau, D., Cho, K., Bengio, Y.: Neural machine translation by jointly
  learning to align and translate.
\newblock arXiv preprint arXiv:1409.0473  (2014)

\bibitem{barth2018distributed}
Barth-Maron, G., Hoffman, M.W., Budden, D., Dabney, W., Horgan, D., Muldal, A.,
  Heess, N., Lillicrap, T.: Distributed distributional deterministic policy
  gradients.
\newblock arXiv preprint arXiv:1804.08617  (2018)

\bibitem{bellemare2017distributional}
Bellemare, M.G., Dabney, W., Munos, R.: A distributional perspective on
  reinforcement learning.
\newblock arXiv preprint arXiv:1707.06887  (2017)

\bibitem{Richard1961Adaptive}
Bellman, R.E.: Adaptive control processes: a guided tour.
\newblock Princeton University Press (1961)

\bibitem{crowninshield1981}
Crowninshield, R.D., Brand, R.A.: A physiologically based criterion of muscle
  force prediction in locomotion.
\newblock Journal of Biomechanics \textbf{14}(11), 793 -- 801 (1981)

\bibitem{dabney2017distributional}
Dabney, W., Rowland, M., Bellemare, M.G., Munos, R.: Distributional
  reinforcement learning with quantile regression.
\newblock arXiv preprint arXiv:1710.10044  (2017)

\bibitem{delp2007opensim}
Delp, S.L., Anderson, F.C., Arnold, A.S., Loan, P., Habib, A., John, C.T.,
  Guendelman, E., Thelen, D.G.: Opensim: open-source software to create and
  analyze dynamic simulations of movement.
\newblock IEEE transactions on biomedical engineering \textbf{54}(11),
  1940--1950 (2007)

\bibitem{baselines}
Dhariwal, P., Hesse, C., Plappert, M., Radford, A., Schulman, J., Sidor, S.,
  Wu, Y.: {OpenAI Baselines}.
\newblock \url{https://github.com/openai/baselines} (2017)

\bibitem{dietterich2000ensemble}
Dietterich, T.G., et~al.: Ensemble methods in machine learning.
\newblock Multiple classifier systems \textbf{1857}, 1--15 (2000)

\bibitem{farris2014exo}
Farris, D.J., Hicks, J.L., Delp, S.L., Sawicki, G.S.: Musculoskeletal modelling
  deconstructs the paradoxical effects of elastic ankle exoskeletons on
  plantar-flexor mechanics and energetics during hopping.
\newblock Journal of Experimental Biology \textbf{217}(22), 4018--4028 (2014)

\bibitem{fortunato2017noisy}
Fortunato, M., Azar, M.G., Piot, B., Menick, J., Osband, I., Graves, A., Mnih,
  V., Munos, R., Hassabis, D., Pietquin, O., et~al.: Noisy networks for
  exploration.
\newblock arXiv preprint arXiv:1706.10295  (2017)

\bibitem{fujimoto2018addressing}
Fujimoto, S., van Hoof, H., Meger, D.: Addressing function approximation error
  in actor-critic methods.
\newblock arXiv preprint arXiv:1802.09477  (2018)

\bibitem{haarnoja2018soft}
Haarnoja, T., Zhou, A., Abbeel, P., Levine, S.: Soft actor-critic: Off-policy
  maximum entropy deep reinforcement learning with a stochastic actor.
\newblock arXiv preprint arXiv:1801.01290  (2018)

\bibitem{horgan2018distributed}
Horgan, D., Quan, J., Budden, D., Barth-Maron, G., Hessel, M., Van~Hasselt, H.,
  Silver, D.: Distributed prioritized experience replay.
\newblock arXiv preprint arXiv:1803.00933  (2018)

\bibitem{huang2017snapshot}
Huang, G., Li, Y., Pleiss, G., Liu, Z., Hopcroft, J.E., Weinberger, K.Q.:
  Snapshot ensembles: Train 1, get m for free.
\newblock arXiv preprint arXiv:1704.00109  (2017)

\bibitem{huang2017learning}
Huang, Z., Zhou, S., Zhuang, B., Zhou, X.: Learning to run with actor-critic
  ensemble.
\newblock arXiv preprint arXiv:1712.08987  (2017)

\bibitem{Ian2016Deep}
Ian~Osband Charles~Blundell, A.P.B.V.R.: Deep exploration via bootstrapped dqn
  (2016)

\bibitem{jaskowski2018rltorunfast}
Ja\'skowski, W., Lykkeb{\o}, O.R., Toklu, N.E., Trifterer, F., Buk, Z.,
  Koutn\'{i}k, J., Gomez, F.: {Reinforcement Learning to Run... Fast}.
\newblock In: S.~Escalera, M.~Weimer (eds.) NIPS 2017 Competition Book.
  Springer, Springer (2018)

\bibitem{john2013stabilisation}
John, C.T., Anderson, F.C., Higginson, J.S., Delp, S.L.: Stabilisation of
  walking by intrinsic muscle properties revealed in a three-dimensional
  muscle-driven simulation.
\newblock Computer methods in biomechanics and biomedical engineering
  \textbf{16}(4), 451--462 (2013)

\bibitem{kidzinski2018learning}
Kidzi{\'n}ski, {\L}., Mohanty, S.P., Ong, C., Huang, Z., Zhou, S., Pechenko,
  A., Stelmaszczyk, A., Jarosik, P., Pavlov, M., Kolesnikov, S., et~al.:
  Learning to run challenge solutions: Adapting reinforcement learning methods
  for neuromusculoskeletal environments.
\newblock arXiv preprint arXiv:1804.00361  (2018)

\bibitem{kidzinski2018learningtorun}
Kidzi\'nski, {\L}., Sharada, M.P., Ong, C., Hicks, J., Francis, S., Levine, S.,
  Salath\'e, M., Delp, S.: Learning to run challenge: Synthesizing
  physiologically accurate motion using deep reinforcement learning.
\newblock In: S.~Escalera, M.~Weimer (eds.) NIPS 2017 Competition Book.
  Springer, Springer (2018)

\bibitem{klambauer2017self}
Klambauer, G., Unterthiner, T., Mayr, A., Hochreiter, S.: Self-normalizing
  neural networks.
\newblock arXiv preprint arXiv:1706.02515  (2017)

\bibitem{lee2017exosuit}
Lee, G., Kim, J., Panizzolo, F., Zhou, Y., Baker, L., Galiana, I., Malcolm, P.,
  Walsh, C.: Reducing the metabolic cost of running with a tethered soft
  exosuit.
\newblock Science Robotics \textbf{2}(6) (2017)

\bibitem{seungjaeryanlee}
Lee, S.R.: {Helper for NIPS 2018: AI for Prosthetics}.
\newblock \url{https://github.com/seungjaeryanlee/osim-rl-helper} (2018)

\bibitem{lillicrap2015continuous}
Lillicrap, T.P., Hunt, J.J., Pritzel, A., Heess, N., Erez, T., Tassa, Y.,
  Silver, D., Wierstra, D.: Continuous control with deep reinforcement
  learning.
\newblock arXiv preprint arXiv:1509.02971  (2015)

\bibitem{loshchilov2017SGDR}
Loshchilov, I., Hutter, F.: Sgdr: Stochastic gradient descent with warm
  restarts.
\newblock In: International Conference on Learning Representations (ICLR) 2017
  Conference Track (2017)

\bibitem{mnih2015human}
Mnih, V., Kavukcuoglu, K., Silver, D., Rusu, A.A., Veness, J., Bellemare, M.G.,
  Graves, A., Riedmiller, M., Fidjeland, A.K., Ostrovski, G., et~al.:
  Human-level control through deep reinforcement learning.
\newblock Nature \textbf{518}(7540), 529--533 (2015)

\bibitem{moritz2018ray}
Moritz, P., Nishihara, R., Wang, S., Tumanov, A., Liaw, R., Liang, E., Elibol,
  M., Yang, Z., Paul, W., Jordan, M.I., et~al.: Ray: A distributed framework
  for emerging $\{$AI$\}$ applications.
\newblock In: 13th $\{$USENIX$\}$ Symposium on Operating Systems Design and
  Implementation ($\{$OSDI$\}$ 18), pp. 561--577 (2018)

\bibitem{ong2017walking}
Ong, C.F., Geijtenbeek, T., Hicks, J.L., Delp, S.L.: Predictive simulations of
  human walking produce realistic cost of transport at a range of speeds.
\newblock In: Proceedings of the 16th International Symposium on Computer
  Simulation in Biomechanics, pp. 19--20 (2017)

\bibitem{pardo2017time}
Pardo, F., Tavakoli, A., Levdik, V., Kormushev, P.: Time limits in
  reinforcement learning.
\newblock arXiv preprint arXiv:1712.00378  (2017)

\bibitem{pavlov}
{Pavlov}, M., {Kolesnikov}, S., {Plis}, S.M.: {Run, skeleton, run: skeletal
  model in a physics-based simulation}.
\newblock ArXiv e-prints  (2017)

\bibitem{peng2018deepmimic}
Peng, X.B., Abbeel, P., Levine, S., van~de Panne, M.: Deepmimic: Example-guided
  deep reinforcement learning of physics-based character skills.
\newblock arXiv preprint arXiv:1804.02717  (2018)

\bibitem{plappert2017parameter}
Plappert, M., Houthooft, R., Dhariwal, P., Sidor, S., Chen, R.Y., Chen, X.,
  Asfour, T., Abbeel, P., Andrychowicz, M.: Parameter space noise for
  exploration.
\newblock arXiv preprint arXiv:1706.01905 (2) (2017)

\bibitem{ross2011reduction}
Ross, S., Gordon, G., Bagnell, D.: A reduction of imitation learning and
  structured prediction to no-regret online learning.
\newblock In: Proceedings of the fourteenth international conference on
  artificial intelligence and statistics, pp. 627--635 (2011)

\bibitem{schaul2015prioritized}
Schaul, T., Quan, J., Antonoglou, I., Silver, D.: Prioritized experience
  replay.
\newblock arXiv preprint arXiv:1511.05952  (2015)

\bibitem{schulman2015trust}
Schulman, J., Levine, S., Abbeel, P., Jordan, M.I., Moritz, P.: Trust region
  policy optimization.
\newblock In: ICML, pp. 1889--1897 (2015)

\bibitem{schulman2015high}
Schulman, J., Moritz, P., Levine, S., Jordan, M., Abbeel, P.: High-dimensional
  continuous control using generalized advantage estimation.
\newblock arXiv preprint arXiv:1506.02438  (2015)

\bibitem{ppo}
Schulman, J., Wolski, F., Dhariwal, P., Radford, A., Klimov, O.: Proximal
  policy optimization algorithms.
\newblock CoRR \textbf{abs/1707.06347} (2017).
\newblock \urlprefix\url{http://arxiv.org/abs/1707.06347}

\bibitem{schulman2017proximal}
Schulman, J., Wolski, F., Dhariwal, P., Radford, A., Klimov, O.: Proximal
  policy optimization algorithms.
\newblock arXiv preprint arXiv:1707.06347  (2017)

\bibitem{seth2018opensim}
Seth, A., Hicks, J., Uchida, T., Habib, A., Dembia, C., Dunne, J., Ong, C.,
  DeMers, M., Rajagopal, A., Millard, M., Hamner, S., Arnold, E., Yong, J.,
  Lakshmikanth, S., Sherman, M., Delp, S.: Opensim: Simulating musculoskeletal
  dynamics and neuromuscular control to study human and animal movement.
\newblock Plos Computational Biology, 14(7).  (2018)

\bibitem{silver2014deterministic}
Silver, D., Lever, G., Heess, N., Degris, T., Wierstra, D., Riedmiller, M.:
  Deterministic policy gradient algorithms.
\newblock In: Proceedings of the 31st International Conference on Machine
  Learning (ICML-14), pp. 387--395 (2014)

\bibitem{song2015neural}
Song, S., Geyer, H.: A neural circuitry that emphasizes spinal feedback
  generates diverse behaviours of human locomotion.
\newblock The Journal of physiology \textbf{593}(16), 3493--3511 (2015)

\bibitem{ivan_sosin_2018_1938263}
Sosin, I., Svidchenko, O., Malysheva, A., Kudenko, D., Shpilman, A.: {Framework
  for Deep Reinforcement Learning with GPU-CPU Multiprocessing} (2018).
\newblock \doi{10.5281/zenodo.1938263}.
\newblock \urlprefix\url{https://doi.org/10.5281/zenodo.1938263}

\bibitem{Sutton1999}
Sutton, R.S., Precup, D., Singh, S.: {Between {MDP}s and semi-{MDP}s: A
  framework for temporal abstraction in reinforcement learning}.
\newblock Artificial Intelligence \textbf{112} (1999)

\bibitem{thelen2003cmc}
Thelen, D.G., Anderson, F.C., Delp, S.L.: Generating dynamic simulations of
  movement using computed muscle control.
\newblock Journal of Biomechanics \textbf{36}(3), 321--328 (2003)

\bibitem{thelen2003generating}
Thelen, D.G., Anderson, F.C., Delp, S.L.: Generating dynamic simulations of
  movement using computed muscle control.
\newblock Journal of biomechanics \textbf{36}(3), 321--328 (2003)

\bibitem{uchida2016device}
Uchida, T.K., Seth, A., Pouya, S., Dembia, C.L., Hicks, J.L., Delp, S.L.:
  Simulating ideal assistive devices to reduce the metabolic cost of running.
\newblock PLOS ONE \textbf{11}(9), 1--19 (2016).
\newblock \doi{10.1371/journal.pone.0163417}

\bibitem{Yuxin2017Training}
Wu, Y., Tian, Y.: Training agent for first-person shooter game with
  actor-critic curriculum learning  (2017)

\bibitem{Yoshua2009Curriculum}
Yoshua, B., J\'erome, L., Ronan, C., Jason, W.: Curriculum learning  (2009)

\end{thebibliography}
\end{document}